\PassOptionsToPackage{table}{xcolor}
\documentclass[sigconf]{acmart}
\usepackage{amsmath}
\usepackage[english]{babel}
\usepackage{algorithm}
\usepackage{algpseudocode}
\usepackage{enumitem}
\usepackage{booktabs}
\usepackage{multirow}
\usepackage{graphicx}
\usepackage{subcaption}
\usepackage{tabularx}
\usepackage{array}
\usepackage{booktabs}
\usepackage{xcolor}
\usepackage{balance}

\usepackage[normalem]{ulem}
\useunder{\uline}{\ul}{}
\renewcommand\footnotetextcopyrightpermission[1]{} 
\renewcommand\footnotetextcopyrightpermission[1]{}

\newcommand{\model}{MV-Thinker}

\AtBeginDocument{}

\setcopyright{acmcopyright}
\copyrightyear{2018}
\acmYear{2018}
\acmDOI{XXXXXXX.XXXXXXX}

\acmPrice{15.00}
\acmISBN{978-1-4503-XXXX-X/18/06}

\begin{document}
\title{Think Globally, Generate Personally: Multi-View Thinking with Adaptive Generation for Generative Recommendation}
\author{Yuxuan Hu}
\affiliation{%
  \institution{City University of Hong Kong}
  \city{Hong Kong}
  \country{China}
}
\email{yuxuanhu7-c@my.cityu.edu.hk}

\author{Yingyi Zhang}
\affiliation{%
  \institution{City University of Hong Kong}
  \city{Hong Kong}
  \country{China}
}
\email{yzhang6375-c@my.cityu.edu.hk}

\author{Wenlin Zhang} 
\affiliation{%
  \institution{City University of Hong Kong}
  \city{Hong Kong}
  \country{China}
}
\email{wl.z@my.cityu.edu.hk}

\author{Chao Zhang} 
\affiliation{%
  \institution{City University of Hong Kong}
  \city{Hong Kong}
  \country{China}
}
\email{czhang2328-c@my.cityu.edu.hk}

\author{Haotian Nan} 
\affiliation{%
  \institution{City University of Hong Kong}
  \city{Hong Kong}
  \country{China}
}
\email{haotianan2-c@my.cityu.edu.hk}

\author{Ziwei Liu} 
\affiliation{%
  \institution{City University of Hong Kong}
  \city{Hong Kong}
  \country{China}
}
\email{lziwei2-c@my.cityu.edu.hk}

\author{Jiancheng Dong} 
\affiliation{%
  \institution{City University of Hong Kong}
  \city{Hong Kong}
  \country{China}
}
\email{jiancdong2-c@my.cityu.edu.hk}

\author{Tongzhou Wu} 
\affiliation{%
  \institution{City University of Hong Kong}
  \city{Hong Kong}
  \country{China}
}
\email{tongzhouw3-c@my.cityu.edu.hk}

\author{Beibei Kong}
\affiliation{%
\department{Platform and Content Group}
  \institution{Tencent}
  \city{Shenzhen}
  \country{China}
}
\email{echokong@tencent.com}

\author{Lei Cheng} 
\affiliation{%
\department{Platform and Content Group}
  \institution{Tencent}
  \city{Shenzhen}
  \country{China}
}
\email{raycheng@tencent.com}

\author{Chengxiang Zhuo} 
\affiliation{%
\department{Platform and Content Group}
  \institution{Tencent}
  \city{Shenzhen}
  \country{China}
}
\email{felixzhuo@tencent.com}

\author{Zang Li} 
\affiliation{%
\department{Platform and Content Group}
  \institution{Tencent}
  \city{Shenzhen}
  \country{China}
}
\email{gavinzli@tencent.com}

\author{Wanyu Wang}
\affiliation{%
  \institution{City University of Hong Kong}
  \city{Hong Kong}
  \country{China}
}
\email{wanyuwang4-c@my.cityu.edu.hk}

\author{Xiangyu Zhao}
\authornote{Corresponding author.}
\affiliation{%
  \institution{City University of Hong Kong}
  \city{Hong Kong}
  \country{China}
}
\email{xianzhao@cityu.edu.hk}
\renewcommand{\shortauthors}{Anonymous Author(s)}

\begin{abstract}
Semantic-ID-based (SID) generative recommendation (GR) represents each item as a sequence of discrete tokens and predicts the next item through autoregressive SID generation. However, existing methods face two challenges. First, \textbf{insufficient shared context modeling for user sequences}: full-history and recent behaviors provide overlapping yet complementary evidence, but modeling them within a single sequence may dilute current cues, whereas processing them through separate streams may fail to coordinate them. 
Second, \textbf{insufficient personalized adaptation in progressive SID decisions}: although decoder states evolve with generated prefixes, existing models typically use a shared output-side transformation across recommendation instances, without explicitly selecting a transformation suited to the current user need.

To address these challenges, we propose \textbf{M}ulti-\textbf{V}iew \textbf{Think}ing with
Adaptive Generation (\textbf{\model{}}), a framework that \emph{Thinks Globally} and \emph{Generates Personally}. First, Multi-View Temporal Thinking uses view-specific thinking queries and readers over overlapping full-history and recent-history views, coordinating both sources into complementary user-conditioned contexts before SID generation. 
Second, Hierarchical Residual Adaptation organizes lightweight adapters into nested residual paths before a shared vocabulary projection, adaptively transforming context-bearing decoder states for SID decisions. 
Extensive experiments on four public datasets demonstrate that \model{} consistently outperforms strong discriminative and generative recommendation baselines across multiple ranking metrics. The implementation code and datasets are available on: \url{https://anonymous.4open.science/r/MV-Thinker}.

\end{abstract}

\ccsdesc[500]{Information systems~Recommender systems}
\keywords{Generative Recommendation, Temporal Preference Modeling, Latent Thinking, Adaptive Generation}
\maketitle

\section{Introduction}
\label{sec:introduction}

Recommender systems~\cite{covington2016deep,zhao2019recommending,deng2025onerec} traditionally handle large item catalogs through multi-stage pipelines consisting of retrieval, coarse ranking, and fine ranking. Generative Recommendation (GR)~\cite{geng2022recommendation,bao2025bi} provides an alternative by representing each item as a sequence of discrete semantic identifier (SID) tokens and formulating next-item prediction as autoregressive generation~\cite{rajput2023recommender,wang2024learnable}. Although the prediction target is a specific item, a user's interaction history may support multiple plausible needs~\cite{tan2021sparse,chang2021sequential}. The full history provides a broad behavioral background, whereas recent interactions highlight the user's current activity~\cite{villatel2018recurrent,liu2018stamp,sun2020go}. Effective recommendation therefore requires not only forming an appropriate preference context from these behavioral signals~\cite{li2017neural}, but also translating that context into an item prediction suited to the current need.

Recent studies~\cite{liao2024llara,wang2024eager} have enhanced both user-sequence modeling and SID generation beyond the basic autoregressive framework. For user-sequence modeling, CARE~\cite{lin2026bringing} progressively refines historical semantics through query-anchored reasoning, while GEMs~\cite{zhou2026gems} introduces behavioral evidence from recent, mid-term, and long-term histories through multiple temporal streams. These methods~\cite{lin2026bringing,zhou2026gems} improve preference representations through semantic refinement and multi-timescale modeling. \emph{However, they still have difficulty coordinating the full history and recent interactions within a shared context, preventing complementary evidence from jointly characterizing the current need.} For SID generation, APAO~\cite{yu2026apao} improves the consistency between teacher-forced training and autoregressive inference through prefix-level optimization, while Latte~\cite{hou2026expressiveness} introduces latent-token-conditioned decoding paths to enhance generation flexibility. These methods improve the generation of target SIDs from the perspectives of training objectives and decoding paths. Nevertheless, \emph{they generally transform and use decoder states in a fixed manner, making it difficult to adapt the SID-prediction context to the current user need.}

\begin{figure}[t]
    \setlength{\abovecaptionskip}{0.2\baselineskip}
    \setlength{\belowcaptionskip}{0.2\baselineskip}
    \centering
    \includegraphics[width=\linewidth]{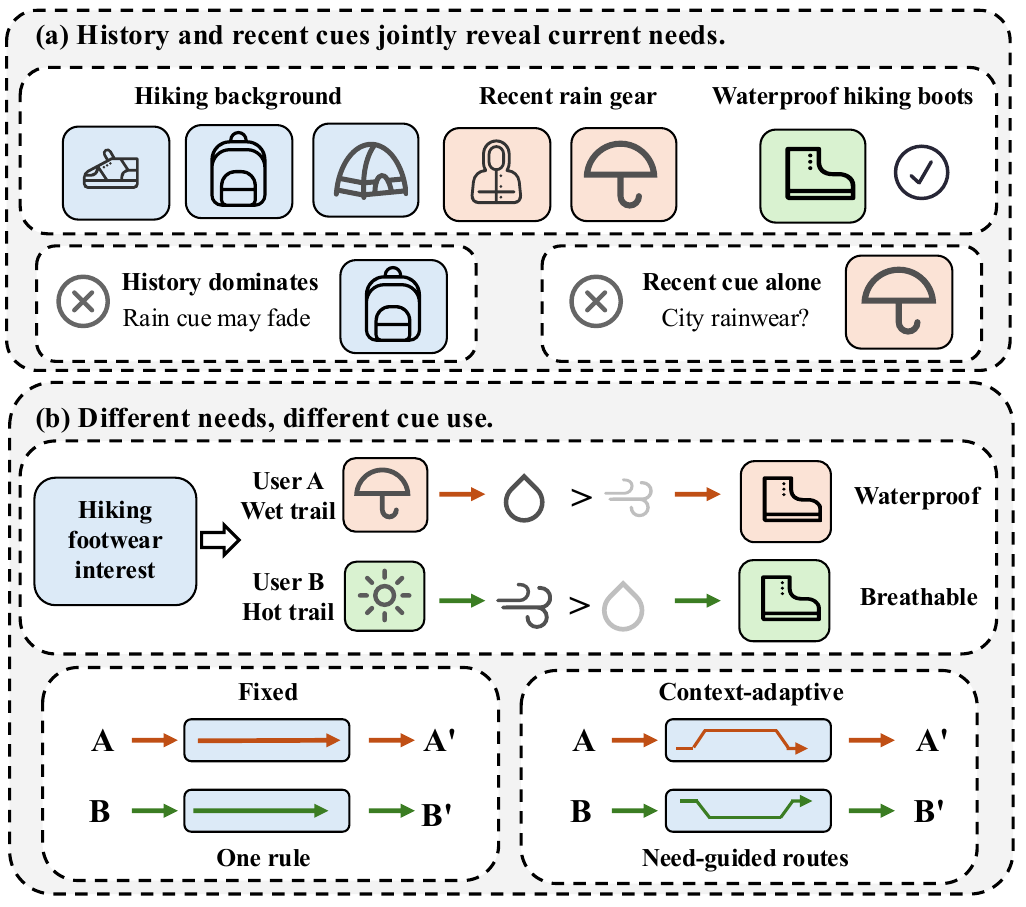}
    \caption{Illustration of two interconnected challenges in SID-based generative recommendation. (a) Insufficient coordination between the full history and recent interactions may either dilute current cues or remove the historical background needed to interpret them. (b) Different recommendation needs may require different uses of contextual cues, whereas a fixed transformation follows the same rule across contexts and lacks need-guided adaptation.}
    \label{fig:introduction}
    \vspace{-10pt}
\end{figure}

Consequently, despite advances in user-sequence modeling and SID generation, existing methods face two interconnected challenges:
\textbf{Challenge 1: \textit{Insufficient shared context modeling for user sequences.}}
The full history provides broad preference context for interpreting user behavior, whereas recent interactions highlight the user's current activity~\cite{villatel2018recurrent,sun2020go}. Modeling the full history as a single sequence may dilute recent cues with long-term behavioral patterns, while processing different temporal scopes through separate streams may inadequately coordinate their overlapping yet complementary evidence. For example, Figure~\ref{fig:introduction}(a) shows that rain-related cues may be diluted by frequent hiking behaviors, while recent activities alone may leave the current need ambiguous. A shared modeling mechanism is therefore needed to preserve immediate cues and the historical background needed to interpret them, forming a user-conditioned context to guide SID generation for the current recommendation instance.
\textbf{Challenge 2: \textit{Insufficient personalized adaptation in progressive SID decisions.}}
Recommendation instances may require different degrees of refinement of context-bearing decoder states. Although these states evolve with generated prefixes, existing methods typically apply a shared output-side transformation across instances, without explicitly selecting one suited to the current recommendation context. As illustrated in Figure~\ref{fig:introduction}(b), users with similar broad interests may have different current needs and therefore require different uses of preference cues. The challenge is to select a suitable transformation per recommendation instance and consistently apply it to evolving decoder states throughout coarse-to-fine SID generation.
Accordingly, our research question is: \emph{How can GR think globally by integrating full-history and recent-history behaviors into a shared preference context, while generating personally through instance-adaptive transformation of decoder states throughout SID generation?}


To answer this question, we propose \textbf{\model{}}, which combines shared user-sequence context modeling with personalized adaptive transformation for SID prediction, enabling the model to \emph{Think Globally, Generate Personally}. To \emph{think globally}, \emph{Multi-View Temporal Thinking} (MVT) constructs overlapping full-history and recent-history views and uses dedicated thinking queries and view-specific readers, whose parameters are shared across users, to form complementary user-conditioned contexts before SID generation. In this way, recent cues are preserved without discarding the broader historical background needed to interpret them. To \emph{generate personally}, \emph{Hierarchical Residual Adaptation} (HRA) organizes lightweight adapters into nested residual paths that transform context-bearing decoder states before the SID vocabulary projection, allowing different recommendation contexts to use suitable transformation depths without maintaining a separate model for each user. 




Our contributions are summarized as follows:
\begin{itemize}[leftmargin=*]
    \item We identify two challenges in GR: insufficient shared context modeling of full and recent behaviors, and insufficient personalized adaptation for SID prediction, motivating the perspective of \emph{Think Globally, Generate Personally}.

    \item We propose \model{}, where MVT organizes full-history and recent-history evidence, and HRA adapts context-bearing decoder states before a shared vocabulary projection.

    \item Extensive experimental results on four public benchmark datasets validate the effectiveness of \model{}.
\end{itemize}

\section{Preliminary}
\label{sec:preliminary}

\subsection{Generative Recommendation}
\label{sec:preliminary_gr}

Let $\mathcal{U}$ and $\mathcal{I}$ denote the sets of users and items. For a user $u\in\mathcal{U}$, the interaction history is $H_u=(i_1,i_2,\ldots,i_n)$ and the next item is $y$. In semantic-ID-based generative recommendation, each item $i\in\mathcal{I}$ is represented by a sequence of discrete tokens, denoted as $S_i=[t_1^i,t_2^i,\ldots,t_L^i]$, where $L$ is the SID length. Given $H_u$, the model generates the SID of $y$ token by token~\cite{rajput2023recommender,wang2024learnable}:
\begin{equation}
    P(S_y\mid H_u)=\prod_{l=1}^{L}P(t_l^y\mid H_u,t_{<l}^y),
    \label{eq:gr_probability}
\end{equation}
where $t_{<l}^y$ denotes the previously generated SID tokens of
the item $y$. Given the training set $\mathcal{D}$, the model is
optimized by the standard token-level cross-entropy:
\begin{equation}
    \mathcal{L}_{\mathrm{GR}}
    =-\sum_{(H_u,y)\in\mathcal{D}}\sum_{l=1}^{L}
    \log P(t_l^y\mid H_u,t_{<l}^y).
    \label{eq:gr_loss}
\end{equation}
During inference, valid-prefix constraints are applied so that each generated SID corresponds to an item in the catalog.

\subsection{Latent Thinking}
\label{sec:preliminary_latent_thinking}

Latent thinking performs intermediate computation in the hidden space to organize preference evidence before item generation~\cite{lin2026bringing}. It commonly uses learnable queries to read the encoded interaction history and construct additional context for the decoder. Unlike explicit chain-of-thought reasoning, the resulting states are continuous representations rather than generated text. In this work, latent thinking specifically refers to query-based preference-context construction before SID generation. It is performed once for the current history and does not introduce iterative reasoning at every SID position. The query parameters are shared across users, while their outputs depend on each user's interaction history.

\section{Method}
\label{sec:method}


\subsection{Overview}
\label{sec:overview}

\begin{figure*}
\setlength\abovecaptionskip{0.2\baselineskip}
\setlength\belowcaptionskip{0.2\baselineskip}
    \centering
    \includegraphics[width=\linewidth]{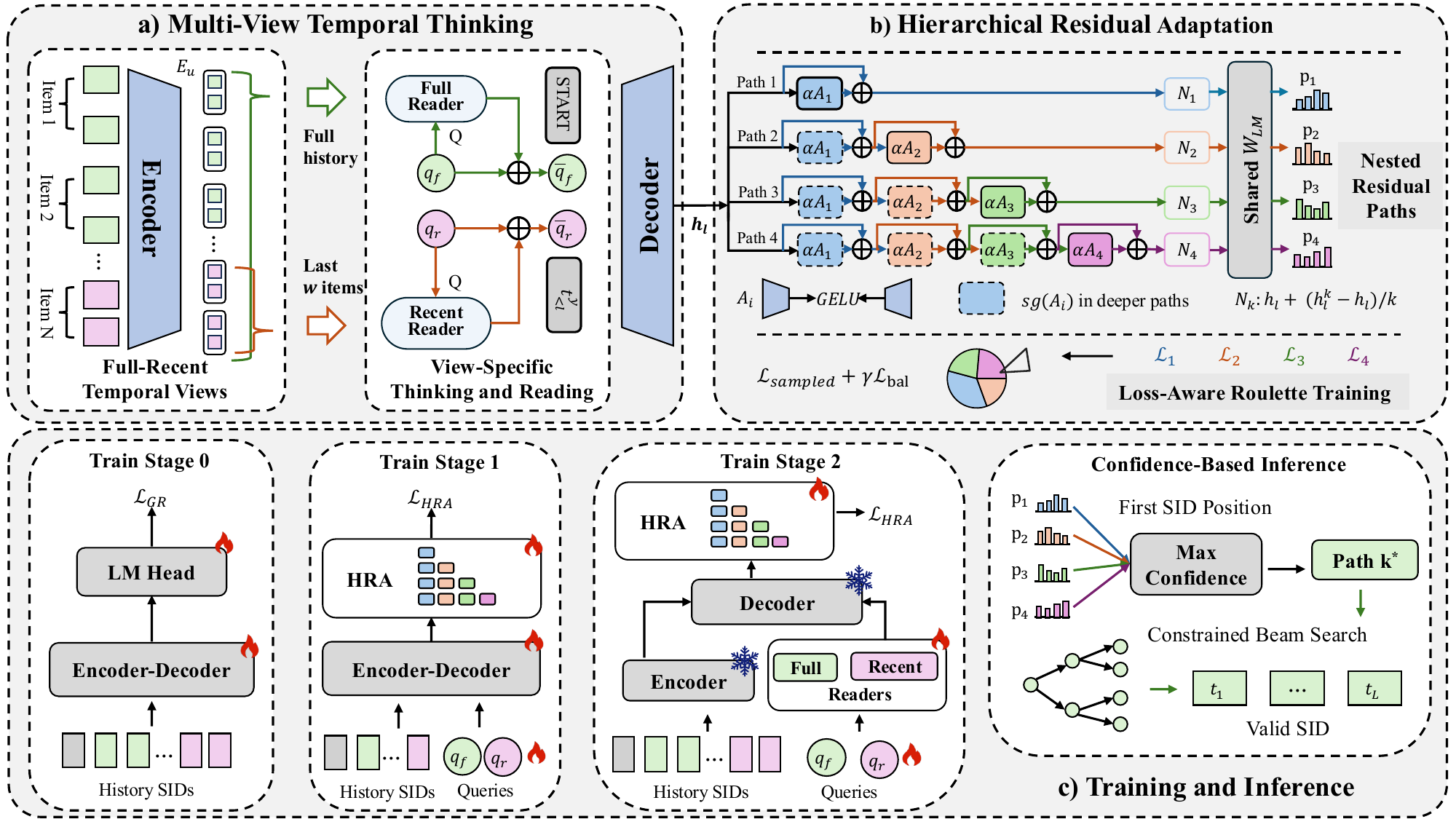}
    \caption{Overview of the proposed \model{} framework.}

    \label{fig:framework}
\end{figure*}

To answer our research question, we propose \model{}, as shown in Figure~\ref{fig:framework}, to integrate full-history and recent-history behaviors into a shared preference context and adapt decoder states for instance-specific progressive SID generation. First, \textbf{Multi-View Temporal Thinking (MVT)} constructs overlapping full-history and recent-history views and forms complementary user-conditioned contexts through view-specific reading before generation. Second, \textbf{Hierarchical Residual Adaptation (HRA)} applies context-dependent residual transformations to decoder states before the shared vocabulary projection, supporting progressive SID decisions at different semantic granularities.


\subsection{Multi-View Temporal Thinking}
\label{sec:thinking}

Modeling a sufficiently informative shared preference context for the current user need is challenging because the full history contains diverse interests, while recent interactions alone provide limited preference background~\cite{zhou2026gems,lin2026bringing}.
To address this challenge, we introduce MVT, which allows the full history to contextualize recent behavior while preserving signals closely related to the current need.
As shown in Figure~\ref{fig:framework}(a), MVT contains two parts: \emph{Full--Recent Temporal Views} defines the full- and recent-history scopes at the item level; \emph{View-Specific Thinking and Reading} assigns a dedicated thinking query to each view and extracts complementary temporal contexts that jointly condition the shared SID generator.

\subsubsection{Full--Recent Temporal Views}

To balance long-term preference patterns with recent preference shifts, we construct two overlapping item-level history views. The full view captures the user's general preference background, while the recent view emphasizes evidence closely related to the current need. The recent view is a suffix of the full history rather than a separate temporal stream. Both views are defined at the item level, so that no item SID is split.

Let $H_u=(i_1,\ldots,i_n)$ be the interaction history. Given a recent-window size $w$, the two views are defined as:
\begin{equation}
    H_u^f=H_u,\qquad
    H_u^r=(i_{\max(1,n-w+1)},\ldots,i_n),
    \label{eq:mvt_views}
\end{equation}
where $f$ and $r$ denote the full and recent views, respectively.
The encoder maps the tokenized history into contextualized hidden states $E_u$.
Accordingly, $E_u^f$ contains all valid history positions, whereas $E_u^r$ contains the positions associated with the last $\min(w,n)$ complete items. The recent view is thus defined at the item level and then mapped to corresponding encoder positions, ensuring that all SID tokens of every selected item are retained.

\subsubsection{View-Specific Thinking and Reading}

Preserving the distinct roles of full- and recent-history evidence within one generator is challenging: a shared reader may conflate the two views, while directly combining their encoder states may let the longer full view dominate recent evidence. To address this issue, we introduce view-specific thinking and reading, pairing each view with a dedicated thinking query and an independent cross-attention reader.

MVT introduces two special tokens, \texttt{<think\_1>} and \texttt{<think\_2>}. Their embeddings $q_f$ and $q_r$ serve as the full-view and recent-view thinking queries, respectively. They are inserted immediately after the decoder start token and before the target SID, forming $[\texttt{START},q_f,q_r,t_{<l}^y]$. The former summarizes the general preference context, while the latter focuses on recent behavioral evidence.

The two queries read their corresponding encoder states through separate multi-head cross-attention readers. For $v\in\{f,r\}$, the reader output and the enhanced query are defined as:
\begin{equation}
z_v=\operatorname{MHA}_v\bigl(\operatorname{LN}(q_v),
\operatorname{LN}(E_u^v),\operatorname{LN}(E_u^v)\bigr),
\qquad \bar q_v=q_v+s z_v,
\label{eq:mvt_readers}
\end{equation}
where $\operatorname{LN}(\cdot)$ denotes layer normalization~\cite{ba2016layer}, and $s$ is a learnable residual scale.
The two independent readers let the full-view query aggregate the user's stable interests and the recent-view query focus on the user's latest behaviors.

At SID position $l$, the decoder state is obtained as:
\begin{equation}
h_l=\operatorname{Dec}\bigl(
E_u,[\texttt{START},\bar q_f,\bar q_r,t_{<l}^y]\bigr).
\label{eq:mvt_decoder}
\end{equation}

\subsection{Hierarchical Residual Adaptation}
\label{sec:adaptation}

Transforming context-bearing decoder states according to the current recommendation context to support personalized progressive SID decisions is challenging.
To address this challenge, we introduce HRA, which provides context-dependent residual transformations at different depths for personalization while retaining a shared backbone and vocabulary projection.
As shown in Figure~\ref{fig:framework}(b), HRA contains three parts: \emph{Nested Residual Paths} builds transformations of increasing depth; \emph{Loss-Aware Roulette Training} encourages path specialization without early collapse; and \emph{Confidence-Based Inference} selects one path and keeps it fixed throughout SID generation.

\subsubsection{Nested Residual Paths}
Different recommendation situations may require different degrees of representation refinement, which a single decoder transformation shared by all instances cannot provide~\cite{elbayad2019depth}. We therefore organize lightweight adapters into nested residual paths, where each deeper path reuses the preceding transformations and adds a further residual update.

Let $A_k$ denote the $k$-th bias-free bottleneck adapter, defined as:
\begin{equation}
A_k(x)=
W_{\mathrm{up}}^k
\operatorname{GELU}(W_{\mathrm{down}}^k x),
\label{eq:hra_adapter}
\end{equation}
where $W_{\mathrm{down}}^k$ and $W_{\mathrm{up}}^k$ are the down- and
up-projection matrices of the $k$-th adapter, respectively.

To provide different degrees of state refinement without learning
independent experts, we organize the adapters into nested paths. A
shallow path makes a small correction to the decoder state, whereas a
deeper path progressively refines it by reusing all shallower
transformations and adding further residual updates. Specifically,
path $k$ contains the ordered adapter sequence $(A_1,\ldots,A_k)$:
path 1 uses only $A_1$, path 2 composes
$A_1\!\rightarrow A_2$, and path 4 uses the complete hierarchy
$A_1\!\rightarrow A_2\!\rightarrow A_3\!\rightarrow A_4$. Starting from $h_l^0=h_l$, path $k$ recursively applies its adapters.
To keep the residual magnitude comparable across different depths, the
accumulated residual is normalized by the path depth as follows:
\begin{equation}
h_l^k=h_l^{k-1}+\alpha A_k\!\left(h_l^{k-1}\right),\qquad
\bar h_l^k=h_l+\tfrac{1}{k}\bigl(h_l^k-h_l\bigr),
\label{eq:hra_recursive}
\end{equation}
where $\alpha$ is a residual scaling coefficient that controls the
contribution of each adapter update.

All paths retain the original vocabulary projection. The token
distribution of path $k$ is defined as:
\begin{equation}
p_l^k=\operatorname{softmax}(W_{\mathrm{LM}}\bar h_l^k),
\label{eq:path-distribution}
\end{equation}
where $W_{\mathrm{LM}}$ is the shared vocabulary projection matrix of
the original decoder. Therefore, all paths predict tokens in the same
semantic vocabulary and differ only in state transformation.

\subsubsection{Loss-Aware Roulette Training}
To select suitable adaptation paths for different instances without premature hard routing, we introduce loss-aware roulette training, which favors better-performing paths while preserving exploration through instance-wise stochastic sampling during training.

Specifically, we adopt roulette-wheel selection~\cite{lipowski2012roulette}, assigning higher selection probabilities to lower-loss paths. During teacher-forced training, the residual paths are applied at every supervised decoder position. Let $\Omega_u$ denote the stage-specific set of supervised positions and $y_{u,l}$ the corresponding label. For user $u$, the sequence loss of path $k$ is defined as:
\begin{equation}
\mathcal{L}_u^k=-\frac{1}{|\Omega_u|}
\sum_{l\in\Omega_u}\log p_l^k(y_{u,l}),
\label{eq:path_loss}
\end{equation}
and
$\boldsymbol{\mathcal{L}}_u=(\mathcal{L}_u^1,\ldots,\mathcal{L}_u^K)$
collects the losses of all $K$ paths. The roulette distribution and
sampled path are defined as:
\begin{equation}
\rho_u=\operatorname{softmax}\!\left(
-\operatorname{sg}(\boldsymbol{\mathcal{L}}_u)/\tau\right),
\qquad k_u\sim\operatorname{Cat}(\rho_u),
\label{eq:roulette}
\end{equation}
where $\operatorname{sg}(\cdot)$ denotes stop-gradient, $\tau$ is the
sampling temperature, $\rho_u$ is the path probability vector, and
$\operatorname{Cat}(\rho_u)$ is the categorical distribution from
which the index $k_u$ is sampled. Although all paths are evaluated
to construct the roulette distribution, only the sampled path supplies
the main cross-entropy update per example.

Because deeper paths reuse shallower adapters, updating all preceding
adapters whenever a deeper path is sampled may interfere with their
learned transformations. We therefore retain the outputs of shallower
adapters during forward computation while detaching them from
back-propagation as follows:
\begin{equation}
\widetilde{A}_i^{(k)}(x)=
\begin{cases}
\operatorname{sg}\!\left(A_i(x)\right), & i<k,\\
A_i(x), & i=k,
\end{cases}
\label{eq:adapter_detach}
\end{equation}
where $\widetilde A_i^{(k)}$ denotes the training-time output of adapter
$i$ under path $k$. Thus, the terminal adapter is optimized on top of
the preserved shallower transformations, maintaining the nested
dependency while giving each depth a distinct optimization role.

Roulette sampling may still favor a small subset of paths early in
training. To prevent such collapse and ensure that all paths receive
sufficient training examples, we recompute the non-detached
probabilities
$\bar\rho_u=\operatorname{softmax}(-\boldsymbol{\mathcal{L}}_u/\tau)$
and add a load-balancing term on their mini-batch average. The overall
objective of loss-aware sampling is as follows:
\begin{equation}
\mathcal{L}_{\mathrm{HRA}}
=\mathcal{L}_{\mathrm{sampled}}+\gamma\,\mathcal{L}_{\mathrm{bal}},
\label{eq:hra_loss}
\end{equation}
with
\begin{equation}
\mathcal{L}_{\mathrm{sampled}}
=\frac{1}{|\mathcal{B}|}\sum_{u\in\mathcal{B}}\mathcal{L}_u^{k_u},
\qquad
\mathcal{L}_{\mathrm{bal}}
=\sum_{k=1}^{K}\left(
\frac{1}{|\mathcal{B}|}\sum_{u\in\mathcal{B}}\bar\rho_u^k-\frac{1}{K}
\right)^2,
\label{eq:hra_terms}
\end{equation}
where $\mathcal{B}$ denotes a mini-batch, $\mathcal{L}_{\mathrm{sampled}}$
is the average loss of the sampled paths, $\mathcal{L}_{\mathrm{bal}}$
penalizes the deviation of the average non-detached path probability $\bar\rho_u^k$ from
the uniform distribution, and $\gamma$ controls the regularization
strength.


\subsubsection{Confidence-Based Inference}
During inference, performing complete beam search separately for every
path introduces substantial decoding overhead~\cite{vijayakumar2016diverse}, while switching paths
across SID positions may lead to inconsistent state transformations~\cite{yu2026apao}.
Therefore, we introduce confidence-based inference, which selects one
path at the first real SID position and fixes it throughout the
subsequent item generation. This maintains path consistency while
avoiding repeated beam search over all paths.

Specifically, we evaluate all $K$ residual paths at the first real SID position and
select the path with the highest predicted next-token log-probability as follows:
\begin{equation}
k_u^*=\arg\max_{k}\max_{t\in\mathcal{V}}\log p_1^k(t),
\label{eq:confidence-routing}
\end{equation}
where $\mathcal{V}$ is the SID vocabulary and $k_u^*$ is the path
selected for user $u$. The selected path remains fixed for all
subsequent SID positions and beam descendants. Accordingly, the score
of candidate item $i$ is accumulated under the same path as follows:
\begin{equation}
P(S_i\mid H_u,k_u^*)
=\prod_{l=1}^{L}p_l^{k_u^*}(t_l^i).
\label{eq:selected_path}
\end{equation}
During decoding, constrained beam search restricts each extension to a valid SID prefix in the item catalog and maps each completed sequence back to its corresponding item.

\subsection{Training and Inference}
\label{sec:training_inference}

To address the challenge that jointly learning all components from scratch may entangle basic SID generation with the specialization of newly introduced modules, we adopt a progressive training strategy that preserves previously learned generative knowledge while gradually adding context construction and adaptive generation. As shown in Figure~\ref{fig:framework}(c), the procedure contains three training stages and one inference process, and Algorithm~\ref{alg:mvthinker} in Appendix~\ref{app:algorithm} summarizes the complete procedure.

\textbf{Stage 0: Base Generative Recommendation.}
We train the encoder--decoder backbone with $\mathcal{L}_{\mathrm{GR}}$ in Equation~\eqref{eq:gr_loss}, establishing a stable mapping from interaction histories to valid item SIDs.

\textbf{Stage 1: Thinking-Query and HRA Learning.}
We insert \texttt{<think\_1>} and \texttt{<think\_2>} before the target SID and introduce HRA. With thinking-token labels unmasked, the model predicts \texttt{<think\_1>}, \texttt{<think\_2>}, the target SID, and \texttt{EOS}, while the backbone, queries, and adapters are jointly optimized by Equation~\eqref{eq:hra_loss}. This stage learns the thinking queries and nested residual
paths before view-specific extraction.

\textbf{Stage 2: Multi-View Temporal Refinement.}
We add the full-view and recent-view readers and freeze the backbone. The thinking tokens remain as the decoder prefix with their labels masked, so the loss covers only the target SID and \texttt{EOS}, while the readers, queries, and adapters are optimized by Equation~\eqref{eq:hra_loss}. This
stage focuses optimization on temporal evidence organization and
output-side adaptation without disturbing the learned SID semantics.

\textbf{Inference.}
The model constructs both temporal views, selects a residual path by Equation~\eqref{eq:confidence-routing}, and generates SIDs by constrained beam search following Equation~\eqref{eq:selected_path}.

\section{Experiments}
\label{sec:experiments}

We conduct experiments on four public datasets to answer the following research questions:
\begin{itemize}[leftmargin=*]
    \item \textbf{RQ1:} What is the performance of \model{} compared with SOTA baseline methods?
    
    \item \textbf{RQ2:} How does \model{} benefit from coordinating full-history and recent-history evidence in a shared context?

    \item \textbf{RQ3:} Do different user contexts favor different transformation depths for progressive SID prediction?

    \item \textbf{RQ4:} How do MVT and HRA contribute to the recommendation performance of \model{}?
    
    \item \textbf{RQ5:} What are the effects of the thinking mechanism, temporal modeling, and the global and recent views?
\end{itemize}

\subsection{Experimental Settings}
\label{sec:experimental_settings}

\subsubsection{Datasets}
\label{subsec:datasets}

We use four real-world recommendation datasets from different domains:
1) \textbf{Instruments}, \textbf{Beauty}, and \textbf{Games} are three
categories of the Amazon review datasets~\cite{ni-etal-2019-justifying},
containing user interactions with musical instruments, beauty products,
and video games, respectively; and 2) \textbf{Yelp}\footnote{\url{https://www.yelp.com/dataset}}
is a popular dataset containing user interactions with local businesses
on the Yelp platform. Following previous generative recommendation
studies~\cite{rajput2023recommender,wang2024learnable}, we discard users and items
with fewer than five interactions and adopt the leave-one-out strategy
for sequential recommendation. For each user, the last interaction is
used for testing, the penultimate interaction for validation, and the
remaining interactions for training. The dataset statistics are summarized in
Table~\ref{tab:dataset_statistics}.

\begin{table}[t]
    \centering
    \caption{Statistics of the processed datasets.}
    \label{tab:dataset_statistics}
    \small
    \begin{tabular}{ccccc}
        \toprule
        Dataset & Users & Items & Interactions & Avg. length \\
        \midrule
        Instruments & 24,772 & 9,922  & 206,153 & 8.32 \\
        Beauty      & 22,363 & 12,101 & 198,502 & 8.88 \\
        Games       & 50,546 & 16,859 & 452,989 & 8.96 \\
        Yelp        & 30,431 & 20,033 & 316,354 & 10.40 \\
        \bottomrule
    \end{tabular}
\end{table}

\begin{table*}[t]
\setlength\abovecaptionskip{0.2\baselineskip}
\setlength\belowcaptionskip{0.2\baselineskip}
\centering
\caption{Overall performance with TIGER and LETTER backbones on four datasets. ``+'' denotes a method applied to the preceding backbone. H@$K$ and N@$K$ denote Hit Ratio and NDCG at $K$, respectively. The best reported results within each backbone group are boldfaced, and shaded rows denote our method. $^{*}$ indicates significant improvements over the strongest baseline in the same backbone group under a two-tailed $t$-test.}
\label{tab:overall_performance}
\begingroup
\setlength{\tabcolsep}{3pt}
\renewcommand{\arraystretch}{1.05}
\newcommand{\mvshade}[1]{\cellcolor{black!8}#1}
\footnotesize
\resizebox{\textwidth}{!}{%
\begin{tabular}{@{}c|l|cccccc|c|l|cccccc@{}}
\toprule
Dataset & \multicolumn{1}{c|}{Method} & H@5 & H@10 & H@20 & N@5 & N@10 & N@20 &
Dataset & \multicolumn{1}{c|}{Method} & H@5 & H@10 & H@20 & N@5 & N@10 & N@20 \\
\midrule
\multirow{18}{*}{Instruments} & Caser & 0.0720 & 0.0856 & 0.1099 & 0.0640 & 0.0683 & 0.0744 & \multirow{18}{*}{Yelp} & Caser & 0.0111 & 0.0205 & 0.0317 & 0.0079 & 0.0109 & 0.0137 \\
 & GRU4Rec & 0.0782 & 0.0925 & 0.1122 & 0.0688 & 0.0734 & 0.0783 & & GRU4Rec & 0.0176 & 0.0258 & 0.0384 & 0.0127 & 0.0133 & 0.0166 \\
 & BERT4Rec & 0.0816 & 0.1040 & 0.1227 & 0.0674 & 0.0745 & 0.0795 & & BERT4Rec & 0.0182 & 0.0285 & 0.0451 & 0.0125 & 0.0158 & 0.0199 \\
 & SASRec & 0.0828 & 0.1019 & 0.1283 & 0.0710 & 0.0772 & 0.0838 & & SASRec & 0.0202 & 0.0313 & 0.0469 & 0.0135 & 0.0171 & 0.0210 \\
\cmidrule(lr){2-8}\cmidrule(lr){10-16}
 & TIGER & 0.0853 & 0.1047 & 0.1293 & 0.0727 & 0.0790 & 0.0852 & & TIGER & 0.0216 & 0.0354 & 0.0531 & 0.0138 & 0.0182 & 0.0227 \\
 & \hspace*{0.8em}+CARE & 0.0859 & 0.1052 & 0.1307 & 0.0731 & 0.0792 & 0.0855 & & \hspace*{0.8em}+CARE & 0.0247 & 0.0359 & 0.0532 & 0.0159 & 0.0201 & 0.0249 \\
 & \hspace*{0.8em}+APAO & 0.0875 & 0.1085 & 0.1297 & 0.0743 & 0.0810 & 0.0869 & & \hspace*{0.8em}+APAO & 0.0222 & 0.0365 & 0.0543 & 0.0144 & 0.0189 & 0.0234 \\
 & \hspace*{0.8em}+Latte & 0.0866 & 0.1066 & 0.1306 & 0.0737 & 0.0801 & 0.0867 & & \hspace*{0.8em}+Latte & 0.0230 & 0.0373 & 0.0566 & 0.0146 & 0.0192 & 0.0244 \\
 & \hspace*{0.8em}+GEMs & 0.0866 & 0.1083 & 0.1296 & 0.0739 & 0.0809 & 0.0866 & & \hspace*{0.8em}+GEMs & 0.0228 & 0.0376 & 0.0557 & 0.0143 & 0.0190 & 0.0236 \\
 & \hspace*{0.8em}+IBA & 0.0861 & 0.1066 & 0.1317 & 0.0738 & 0.0805 & 0.0872 & & \hspace*{0.8em}+IBA & 0.0241 & 0.0383 & 0.0567 & 0.0155 & 0.0204 & 0.0247 \\
 & \mvshade{\hspace*{0.8em}+\model{}} & \mvshade{\textbf{0.0882}$^{*}$} & \mvshade{\textbf{0.1088}$^{*}$} & \mvshade{\textbf{0.1325}$^{*}$} & \mvshade{\textbf{0.0751}$^{*}$} & \mvshade{\textbf{0.0817}$^{*}$} & \mvshade{\textbf{0.0877}$^{*}$} & & \mvshade{\hspace*{0.8em}+\model{}} & \mvshade{\textbf{0.0251}$^{*}$} & \mvshade{\textbf{0.0387}$^{*}$} & \mvshade{\textbf{0.0579}$^{*}$} & \mvshade{\textbf{0.0163}$^{*}$} & \mvshade{\textbf{0.0207}$^{*}$} & \mvshade{\textbf{0.0255}$^{*}$} \\
\cmidrule(lr){2-8}\cmidrule(lr){10-16}
 & LETTER & 0.0854 & 0.1049 & 0.1291 & 0.0762 & 0.0827 & 0.0893 & & LETTER & 0.0220 & 0.0358 & 0.0544 & 0.0143 & 0.0186 & 0.0233 \\
 & \hspace*{0.8em}+CARE & 0.0860 & 0.1054 & 0.1298 & 0.0766 & 0.0833 & 0.0892 & & \hspace*{0.8em}+CARE & 0.0223 & 0.0362 & 0.0548 & 0.0144 & 0.0188 & 0.0235 \\
 & \hspace*{0.8em}+APAO & 0.0881 & 0.1119 & 0.1332 & 0.0777 & 0.0856 & 0.0912 & & \hspace*{0.8em}+APAO & 0.0239 & 0.0371 & 0.0553 & 0.0156 & 0.0195 & 0.0242 \\
 & \hspace*{0.8em}+Latte & 0.0867 & 0.1100 & 0.1327 & 0.0760 & 0.0837 & 0.0897 & & \hspace*{0.8em}+Latte & 0.0236 & 0.0388 & 0.0585 & 0.0154 & 0.0199 & 0.0249 \\
 & \hspace*{0.8em}+GEMs & 0.0882 & 0.1115 & 0.1352 & 0.0772 & 0.0849 & 0.0912 & & \hspace*{0.8em}+GEMs & 0.0238 & 0.0377 & 0.0565 & 0.0158 & 0.0198 & 0.0247 \\
 & \hspace*{0.8em}+IBA & 0.0875 & 0.1102 & 0.1369 & 0.0779 & 0.0854 & 0.0924 & & \hspace*{0.8em}+IBA & 0.0228 & 0.0375 & 0.0589 & 0.0150 & 0.0194 & 0.0253 \\
 & \mvshade{\hspace*{0.8em}+\model{}} & \mvshade{\textbf{0.0906}$^{*}$} & \mvshade{\textbf{0.1122}$^{*}$} & \mvshade{\textbf{0.1404}$^{*}$} & \mvshade{\textbf{0.0801}$^{*}$} & \mvshade{\textbf{0.0873}$^{*}$} & \mvshade{\textbf{0.0947}$^{*}$} & & \mvshade{\hspace*{0.8em}+\model{}} & \mvshade{\textbf{0.0250}$^{*}$} & \mvshade{\textbf{0.0391}$^{*}$} & \mvshade{\textbf{0.0595}$^{*}$} & \mvshade{\textbf{0.0164}$^{*}$} & \mvshade{\textbf{0.0208}$^{*}$} & \mvshade{\textbf{0.0260}$^{*}$} \\
\midrule
\multirow{18}{*}{Beauty} & Caser & 0.0235 & 0.0353 & 0.0557 & 0.0144 & 0.0183 & 0.0234 & \multirow{18}{*}{Games} & Caser & 0.0401 & 0.0683 & 0.0985 & 0.0258 & 0.0349 & 0.0423 \\
 & GRU4Rec & 0.0303 & 0.0441 & 0.0657 & 0.0210 & 0.0254 & 0.0308 & & GRU4Rec & 0.0454 & 0.0719 & 0.1117 & 0.0309 & 0.0394 & 0.0495 \\
 & BERT4Rec & 0.0341 & 0.0515 & 0.0733 & 0.0239 & 0.0295 & 0.0350 & & BERT4Rec & 0.0490 & 0.0801 & 0.1228 & 0.0293 & 0.0393 & 0.0500 \\
 & SASRec & 0.0362 & 0.0523 & 0.0732 & 0.0234 & 0.0287 & 0.0339 & & SASRec & 0.0507 & 0.0800 & 0.1193 & 0.0343 & 0.0437 & 0.0536 \\
\cmidrule(lr){2-8}\cmidrule(lr){10-16}
 & TIGER & 0.0380 & 0.0575 & 0.0858 & 0.0246 & 0.0309 & 0.0380 & & TIGER & 0.0533 & 0.0825 & 0.1230 & 0.0356 & 0.0450 & 0.0559 \\
 & \hspace*{0.8em}+CARE & 0.0381 & 0.0604 & 0.0883 & 0.0247 & 0.0324 & 0.0394 & & \hspace*{0.8em}+CARE & 0.0534 & 0.0831 & 0.1309 & 0.0357 & 0.0456 & 0.0564 \\
 & \hspace*{0.8em}+APAO & 0.0397 & 0.0590 & 0.0865 & 0.0264 & 0.0334 & 0.0408 & & \hspace*{0.8em}+APAO & 0.0544 & 0.0862 & 0.1286 & 0.0370 & 0.0469 & 0.0576 \\
 & \hspace*{0.8em}+Latte & 0.0390 & 0.0572 & 0.0863 & 0.0256 & 0.0322 & 0.0398 & & \hspace*{0.8em}+Latte & 0.0541 & 0.0850 & 0.1279 & 0.0365 & 0.0464 & 0.0572 \\
 & \hspace*{0.8em}+GEMs & 0.0401 & 0.0602 & 0.0867 & 0.0266 & 0.0337 & 0.0409 & & \hspace*{0.8em}+GEMs & 0.0548 & 0.0876 & 0.1293 & 0.0374 & 0.0480 & 0.0585 \\
 & \hspace*{0.8em}+IBA & 0.0387 & 0.0593 & 0.0881 & 0.0252 & 0.0326 & 0.0404 & & \hspace*{0.8em}+IBA & 0.0534 & 0.0861 & 0.1297 & 0.0358 & 0.0462 & 0.0580 \\
 & \mvshade{\hspace*{0.8em}+\model{}} & \mvshade{\textbf{0.0405}$^{*}$} & \mvshade{\textbf{0.0607}$^{*}$} & \mvshade{\textbf{0.0887}$^{*}$} & \mvshade{\textbf{0.0275}$^{*}$} & \mvshade{\textbf{0.0340}$^{*}$} & \mvshade{\textbf{0.0411}$^{*}$} & & \mvshade{\hspace*{0.8em}+\model{}} & \mvshade{\textbf{0.0557}$^{*}$} & \mvshade{\textbf{0.0884}$^{*}$} & \mvshade{\textbf{0.1319}$^{*}$} & \mvshade{\textbf{0.0379}$^{*}$} & \mvshade{\textbf{0.0484}$^{*}$} & \mvshade{\textbf{0.0594}$^{*}$} \\
\cmidrule(lr){2-8}\cmidrule(lr){10-16}
 & LETTER & 0.0379 & 0.0594 & 0.0849 & 0.0248 & 0.0318 & 0.0382 & & LETTER & 0.0541 & 0.0839 & 0.1274 & 0.0364 & 0.0460 & 0.0570 \\
 & \hspace*{0.8em}+CARE & 0.0381 & 0.0604 & 0.0884 & 0.0251 & 0.0328 & 0.0392 & & \hspace*{0.8em}+CARE & 0.0544 & 0.0840 & 0.1274 & 0.0364 & 0.0463 & 0.0578 \\
 & \hspace*{0.8em}+APAO & 0.0400 & 0.0633 & 0.0912 & 0.0262 & 0.0337 & 0.0408 & & \hspace*{0.8em}+APAO & 0.0548 & 0.0869 & 0.1278 & 0.0365 & 0.0471 & 0.0582 \\
 & \hspace*{0.8em}+Latte & 0.0410 & 0.0613 & 0.0904 & 0.0271 & 0.0336 & 0.0409 & & \hspace*{0.8em}+Latte & 0.0553 & 0.0868 & 0.1292 & 0.0367 & 0.0470 & 0.0585 \\
 & \hspace*{0.8em}+GEMs & 0.0426 & 0.0651 & 0.0945 & 0.0283 & 0.0356 & 0.0430 & & \hspace*{0.8em}+GEMs & 0.0565 & 0.0879 & 0.1302 & 0.0376 & 0.0479 & 0.0593 \\
 & \hspace*{0.8em}+IBA & 0.0387 & 0.0627 & 0.0892 & 0.0258 & 0.0324 & 0.0394 & & \hspace*{0.8em}+IBA & 0.0559 & 0.0866 & 0.1310 & 0.0372 & 0.0473 & 0.0594 \\
 & \mvshade{\hspace*{0.8em}+\model{}} & \mvshade{\textbf{0.0436}$^{*}$} & \mvshade{\textbf{0.0665}$^{*}$} & \mvshade{\textbf{0.0964}$^{*}$} & \mvshade{\textbf{0.0291}$^{*}$} & \mvshade{\textbf{0.0366}$^{*}$} & \mvshade{\textbf{0.0441}$^{*}$} & & \mvshade{\hspace*{0.8em}+\model{}} & \mvshade{\textbf{0.0570}$^{*}$} & \mvshade{\textbf{0.0885}$^{*}$} & \mvshade{\textbf{0.1315}$^{*}$} & \mvshade{\textbf{0.0386}$^{*}$} & \mvshade{\textbf{0.0488}$^{*}$} & \mvshade{\textbf{0.0596}$^{*}$} \\
\bottomrule
\end{tabular}%
}
\endgroup
\end{table*}

\subsubsection{Evaluation Metrics}
\label{subsec:evaluation_metrics}

Following previous works~\cite{rajput2023recommender,wang2024learnable,lin2026bringing},
we adopt two standard Top-$k$ recommendation metrics, including Hit
Ratio (H@$k$) and Normalized Discounted Cumulative Gain (N@$k$), with
$k\in\{5,10,20\}$. H@$k$ measures whether the ground-truth item appears
among the top-$k$ recommendations, while N@$k$ further considers its
ranking position in the recommendation list. Higher H@$k$ and N@$k$
indicate better recommendation performance.

\subsubsection{Baselines}
\label{subsec:baselines}

To evaluate the effectiveness of \model{}, we compare it with
representative baseline methods, which can be divided into two
categories: discriminative recommendation methods and generative
recommendation methods.

\textit{Discriminative Recommendation.}
These methods learn user preferences directly from historical
interaction sequences and predict the next item through conventional
ranking objectives.
\begin{itemize}[leftmargin=*]
    \item \textbf{Caser}~\cite{tang2018personalized} applies convolutional
    filters to capture local and sequential patterns in recent user
    interactions.
    
    \item \textbf{GRU4Rec}~\cite{hidasi2015session} models user behavior
    sequences with gated recurrent units to capture evolving user
    preferences.
    
    \item \textbf{BERT4Rec}~\cite{sun2019bert4rec} employs bidirectional
self-attention and masked-item prediction to learn contextual
dependencies among items in user behavior sequences.

    \item \textbf{SASRec}~\cite{kang2018self} adopts unidirectional
    self-attention to capture sequential dependencies in a user's
    interaction history.
\end{itemize}

\textit{Generative Recommendation.}
These methods represent items with semantic identifiers and
formulate next-item recommendation as autoregressive sequence
generation.
\begin{itemize}[leftmargin=*]
    \item \textbf{TIGER}~\cite{rajput2023recommender} employs RQ-VAE to
    construct semantic IDs and uses an encoder--decoder model to
    autoregressively generate the SID of the next item.
    
    \item \textbf{LETTER}~\cite{wang2024learnable} incorporates
collaborative signals and diversity regularization for
recommendation-oriented item tokenization.

    \item \textbf{CARE}~\cite{lin2026bringing} introduces cascaded latent
    reasoning to refine the preference context before SID generation,
    making it closely related to our latent thinking design.
    
    \item \textbf{GEMs}~\cite{zhou2026gems} captures user preferences at
    recent, mid-term, and lifecycle scales through separate generative
    streams.
    
    \item \textbf{APAO}~\cite{yu2026apao} adaptively optimizes SID prefixes
    to reduce the discrepancy between training and
    autoregressive generation.
    
    \item \textbf{Latte}~\cite{hou2026expressiveness} conditions SID generation on
    latent tokens and provides multiple decoding paths.
    
    \item \textbf{IBA}~\cite{yang2026reasoning} adaptively allocates latent
    refinement steps to SID positions according to their information gain,
    providing position-dependent computation during generation.

\end{itemize}

\subsubsection{Implementation Details}
\label{subsec:implementation_details}

All experiments are repeated with three random seeds, $\{42,43,44\}$,
and the average results are reported on 8 NVIDIA RTX PRO 6000 GPUs. We use the SIDs
generated by the RQ-VAE-based tokenization pipeline and adopt a
T5-style~\cite{raffel2020exploring} encoder--decoder backbone with a hidden
dimension of 128 and four Transformer layers~\cite{vaswani2017attention}.
For MVT, we set the recent-window size to $w=5$ items. HRA contains four nested
residual paths with a bottleneck dimension of 32, a residual scaling
coefficient $\alpha=0.2$, and a roulette temperature $\tau=1.0$. We
optimize the model using AdamW~\cite{loshchilov2017decoupled} with a batch size of 256 and two gradient
accumulation steps. The learning rates are set to $5\times10^{-4}$ for
Stages~0 and~1 and $2\times10^{-4}$ for Stage~2, with a weight decay of
$0.01$, a warm-up ratio of $0.01$, and cosine learning-rate decay. The training epochs is 200. During inference, we use
constrained beam search with a beam size of 20.

\subsection{Overall Performance (RQ1)}
\label{sec:overall_performance}


To evaluate whether \model{} outperforms state-of-the-art recommendation methods, we compare it with discriminative and generative baselines on four datasets under both TIGER and LETTER backbones (Table~\ref{tab:overall_performance}). We can find:
\textbf{1) \model{} achieves the best results on all metrics under both backbones}, demonstrating consistent gains in both hit accuracy and ranking quality. 
\textbf{2) Both generative backbones outperform the strongest discriminative baseline}, demonstrating the effectiveness of SID-based generation in these settings. \model{} further improves both backbones, suggesting that its gains extend across the two tested item-tokenization schemes.
\textbf{3) The strongest baseline varies by dataset, backbone, and metric}: sequence-side methods (CARE and GEMs) are often competitive on Beauty and Games, while generator-side methods (APAO, Latte, and IBA) frequently lead on Instruments, with metric-specific exceptions. \model{} consistently outperforms both
types of enhancement, supporting its joint design of coordinating full-history and recent-history evidence through MVT and adapting SID prediction through HRA. 

\begin{table}[t]
\setlength\abovecaptionskip{0.2\baselineskip}
\setlength\belowcaptionskip{0.2\baselineskip}
    \centering
    \caption{SID catalog coverage (\%) on Instruments under
    different recommendation budgets $K$.}
    \label{tab:catalog_coverage}
    \small
    \setlength{\tabcolsep}{5pt}
    \begin{tabular}{lccccc}
        \toprule
        Model & $K=1$ & $K=3$ & $K=5$ & $K=10$ & $K=20$ \\
        \midrule
        TIGER
        & 4.31 & 7.15 & 9.27 & 13.05 & 19.56 \\
        MV-Thinker
        & \textbf{5.51}
        & \textbf{9.75}
        & \textbf{12.13}
        & \textbf{17.59}
        & \textbf{25.50} \\
        \bottomrule
    \end{tabular}
\end{table}


\begin{figure}[t]
\setlength\abovecaptionskip{0.2\baselineskip}
\setlength\belowcaptionskip{0.2\baselineskip}
    \centering
    \includegraphics[width=\linewidth]{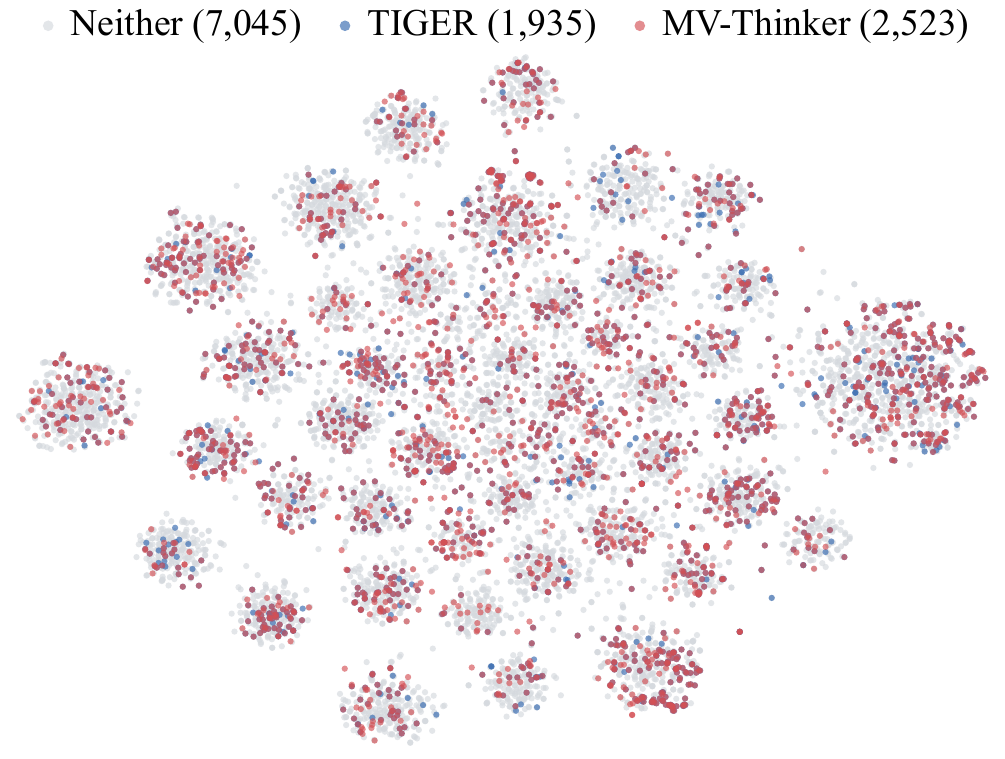}
    \caption{Visualization of Top-20 SID coverage on Instruments.
Each point represents a unique SID on a shared t-SNE map.
Gray points are uncovered, while blue and red markers indicate
coverage by TIGER and MV-Thinker, respectively.}

    \label{fig:recommendation_coverage}
    \vspace{-10pt}
\end{figure}



\subsection{Shared Context Modeling (RQ2)}
\label{sec:coverage_analysis}


To verify whether \model{} can sufficiently model shared context  for user sequences, we compare the recommendation behavior of \model{} and TIGER on Instruments through SID catalog coverage (Table~\ref{tab:catalog_coverage}) and coverage visualization (Figure~\ref{fig:recommendation_coverage}).
We aggregate each model's recommended SIDs across all test users and count distinct codes, defining coverage as:
\begin{equation}
\operatorname{Coverage}_{M}@K =
\frac{
    \left|
    \bigcup_{u \in \mathcal{U}} \mathcal{R}_{M}^{K}(u)
    \right|
}{
    |\mathcal{C}|
}
\times 100\%,
\label{eq:catalog_coverage}
\end{equation}
where $\mathcal{U}$ denotes the evaluated test users, $\mathcal{C}$
denotes all unique catalog SIDs, and $\mathcal{R}_{M}^{K}(u)$
denotes the top-$K$ SIDs recommended by model $M$ to user $u$.


As shown in Table~\ref{tab:catalog_coverage}, we can find:
\textbf{1)\model{} achieves higher coverage at every cutoff}, increasing coverage from 19.56\% to 25.50\% at $K=20$. Figure~\ref{fig:recommendation_coverage} offers a complementary view of the covered SIDs on a shared catalog map. Together with the improved Hit@20 and NDCG@20 in Table~\ref{tab:overall_performance}, these results show broader catalog exposure alongside better target ranking under the same recommendation budget. 
\textbf{2) \model{} broadens catalog coverage while improving ranking accuracy}, exposing more distinct SIDs without an accuracy trade-off. Its shared context combines recent behavioral cues with the historical background needed to interpret them, providing coordinated temporal evidence for SID generation. The joint gains suggest that this design helps accommodate heterogeneous user needs through broader yet relevant recommendations, alleviating the limitations of insufficient coordination between full-history and recent-history evidence.

\subsection{Personalized Context Adaptation (RQ3)}
\label{sec:hra_head}

\begin{figure}[t]
\centering
\includegraphics[width=\linewidth]
{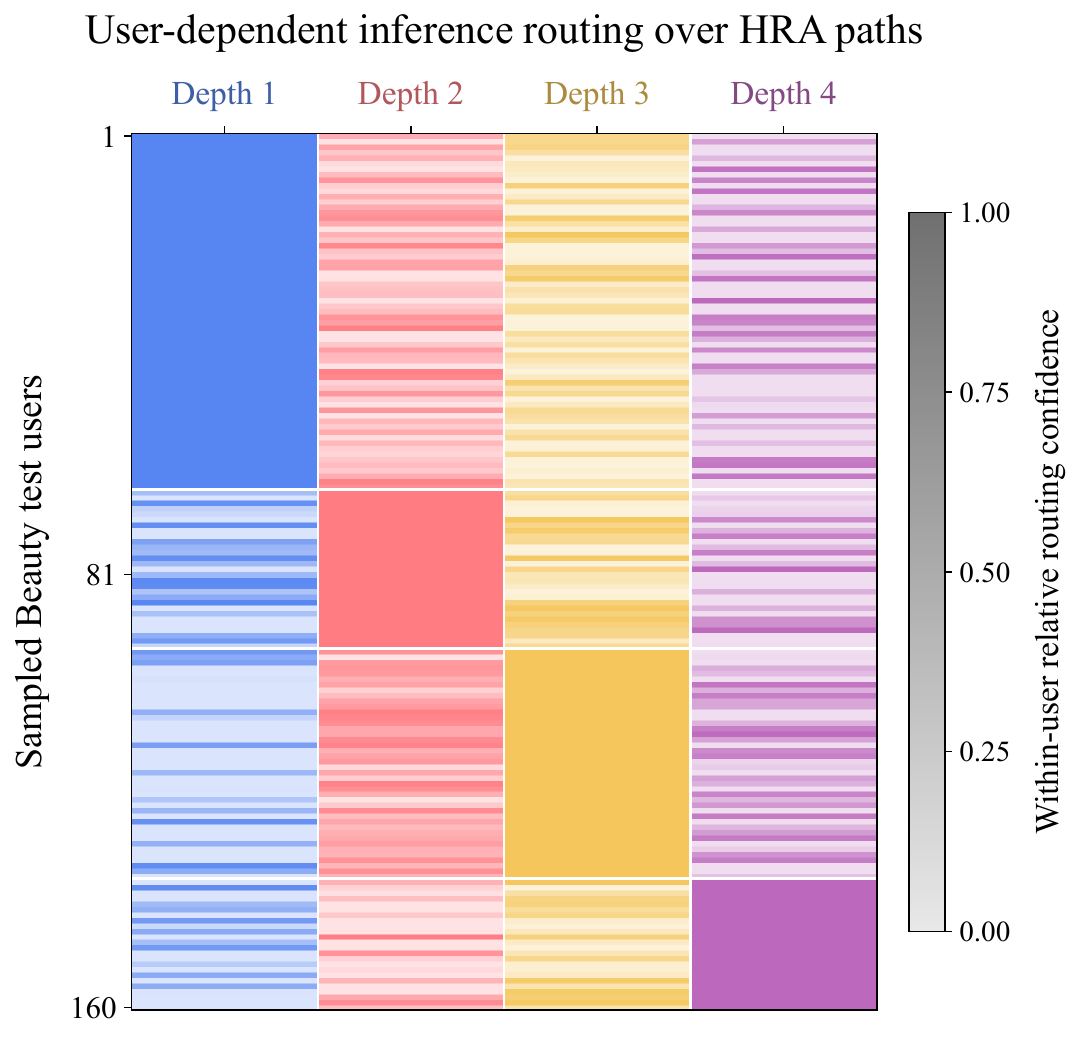}
\caption{User-dependent inference routing over HRA paths on Beauty.
Rows represent sampled test users, and columns indicate
residual depths. Scores are first-position log confidences,
min--max normalized per user. Darker cells indicate higher
relative confidence, and users are grouped by the path
selected during inference.}
\vspace{-10pt}
    \label{fig:beauty-path-preference}
\end{figure}

To verify whether \model{} can adapt personally in progressive SID decisions, we examine its actual routing decisions over HRA paths. For 160 uniformly sampled Beauty test users, we compute each path's log confidence, defined as the log of its maximum next-token probability at the first SID prediction position, using only the observed history. As shown in Figure~\ref{fig:beauty-path-preference}, which presents within-user min--max-normalized scores with users grouped by their selected path, we find that \textbf{HRA learns complementary context transformations suited to different user sequences}: depths~1--4 are selected for 40.6\%, 18.1\%, 26.3\%, and 15.0\% of users, respectively. The selected path remains fixed throughout SID generation for each recommendation instance. Its hierarchical paths thus provide distinct refinement options for personalized SID prediction, alleviating the restriction of applying a single fixed output-side transformation across heterogeneous user contexts.

\subsection{Ablation Study (RQ4)}
\label{sec:ablation_study}

\begin{figure}[t]
\centering
\begin{subfigure}[t]{0.49\columnwidth}
\centering
\includegraphics[width=\linewidth]{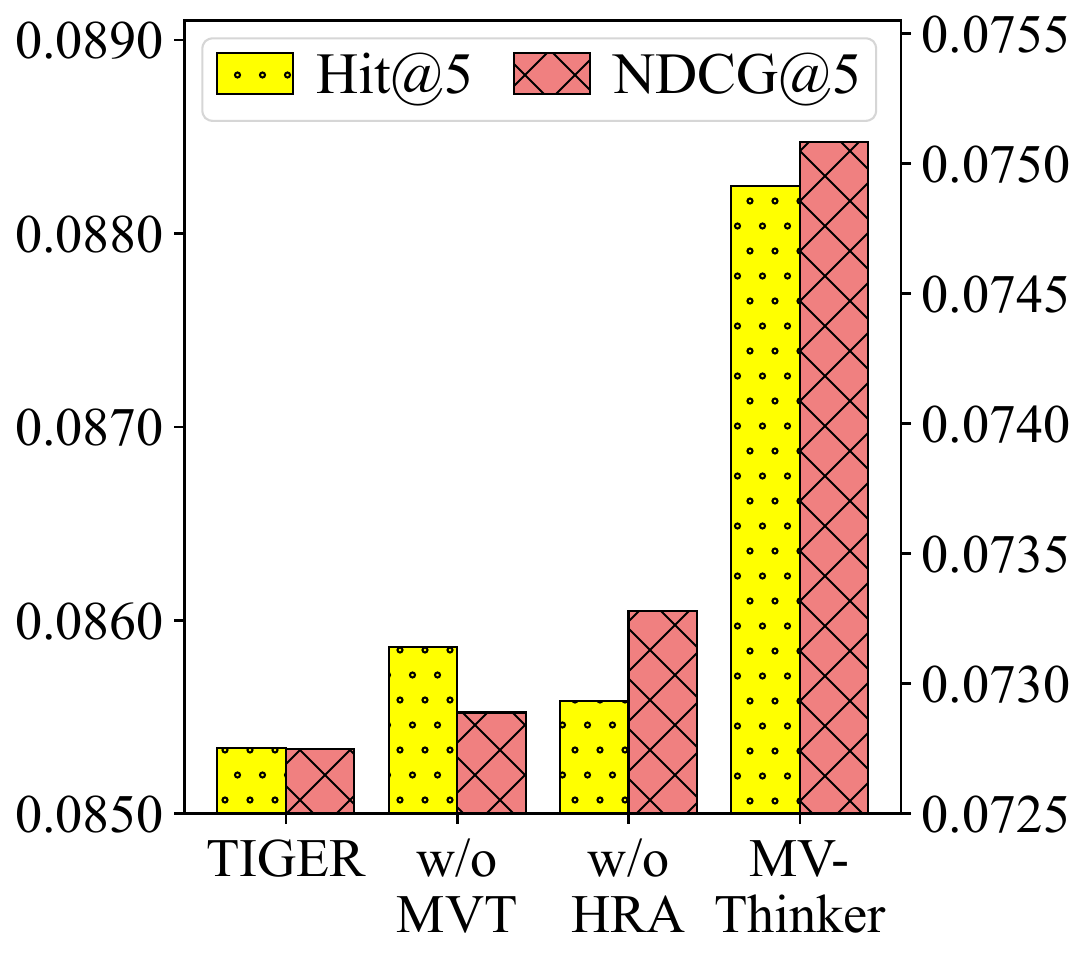}
\caption{Instruments}
\label{fig:ablation_instrument}
\end{subfigure}%
\hfill
\begin{subfigure}[t]{0.49\columnwidth}
\centering
\includegraphics[width=\linewidth]{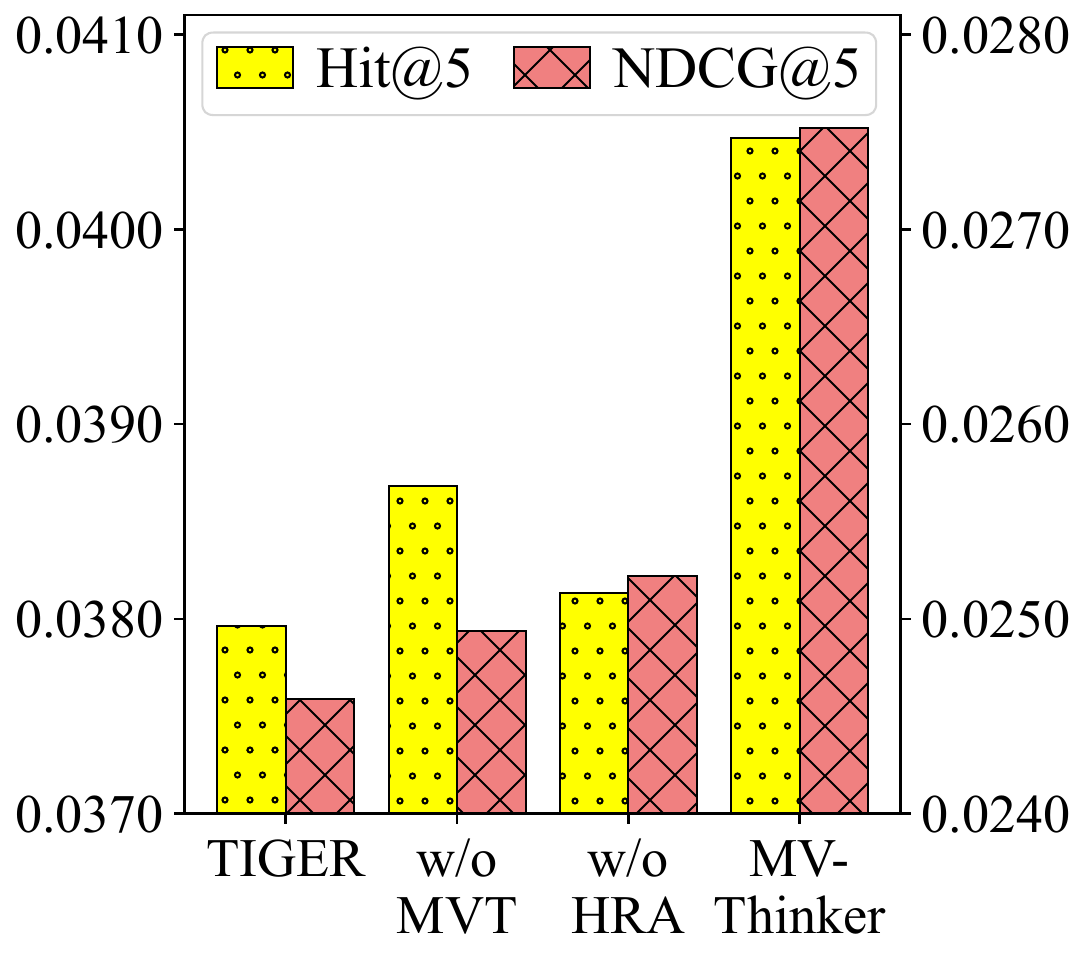}
\caption{Beauty}
\label{fig:ablation_beauty}
\end{subfigure}
\caption{Ablation study of \model{} on the Instruments and Beauty datasets. ``w/o MVT'' and ``w/o HRA'' denote the variants without Multi-View Temporal Thinking and Hierarchical Residual Adaptation, respectively.}
\label{fig:ablation}
\end{figure}

To investigate how MVT and HRA contribute to recommendation
performance, we conduct an ablation study on Instruments and
Beauty by separately removing each module from the framework.
As shown in Figure~\ref{fig:ablation}(a--b), removing either
module reduces Hit@5 and NDCG@5, while both single-module
variants still outperform TIGER. Removing HRA causes a larger
drop in Hit@5, whereas removing MVT causes a larger drop in
NDCG@5 on both datasets. These observations indicate that the
two modules contribute differently to recommendation performance.
Overall, \textbf{both MVT and HRA improve recommendation quality,
and combining them achieves better performance than either
module alone on both datasets, demonstrating their complementary
contributions to the gains of \model{}.}

\begin{figure}[t]
\centering
\begin{subfigure}[t]{0.49\columnwidth}
\centering
\includegraphics[width=\linewidth]{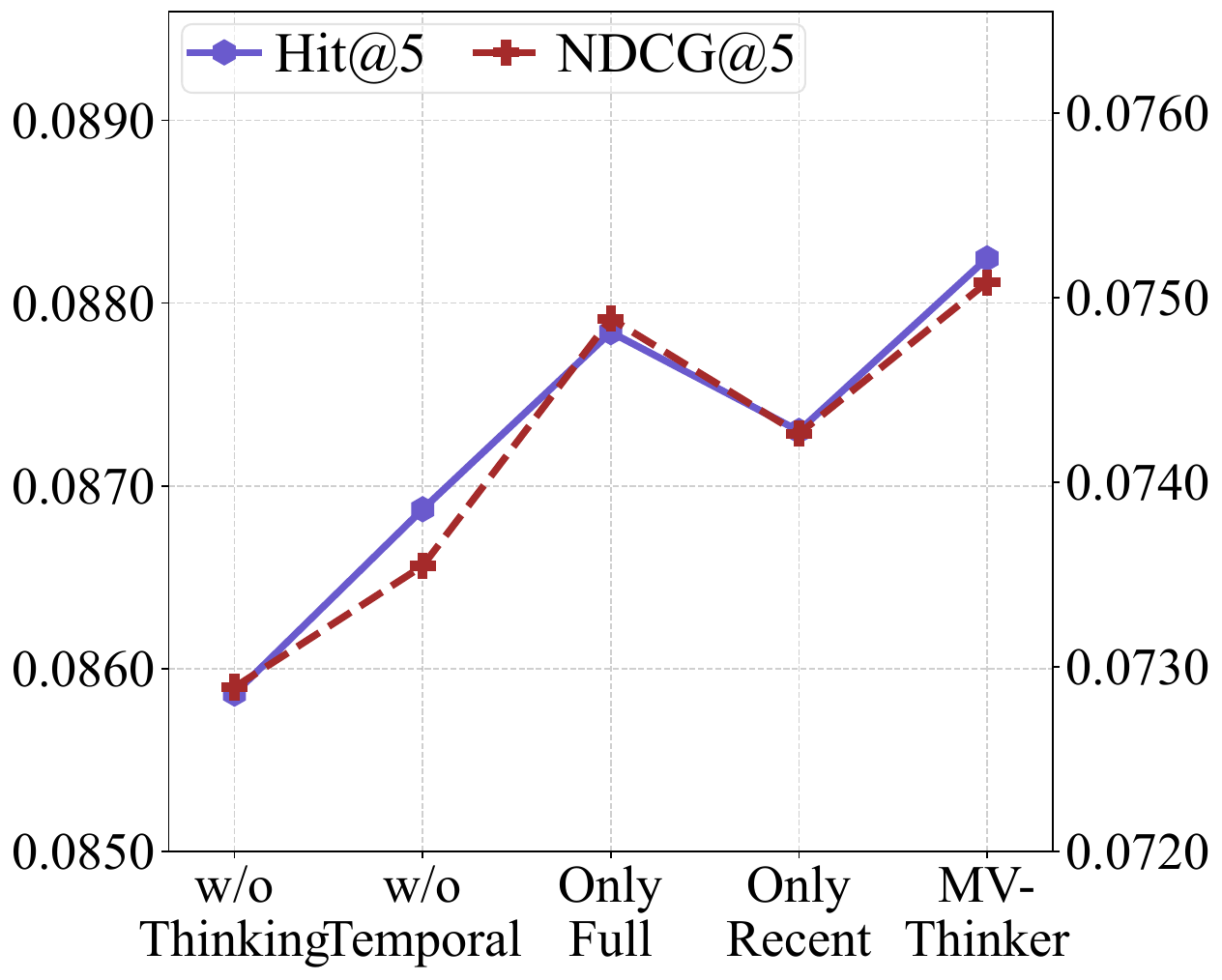}
\caption{Instruments}
\label{fig:component_instrument}
\end{subfigure}%
\hfill
\begin{subfigure}[t]{0.49\columnwidth}
\centering
\includegraphics[width=\linewidth]{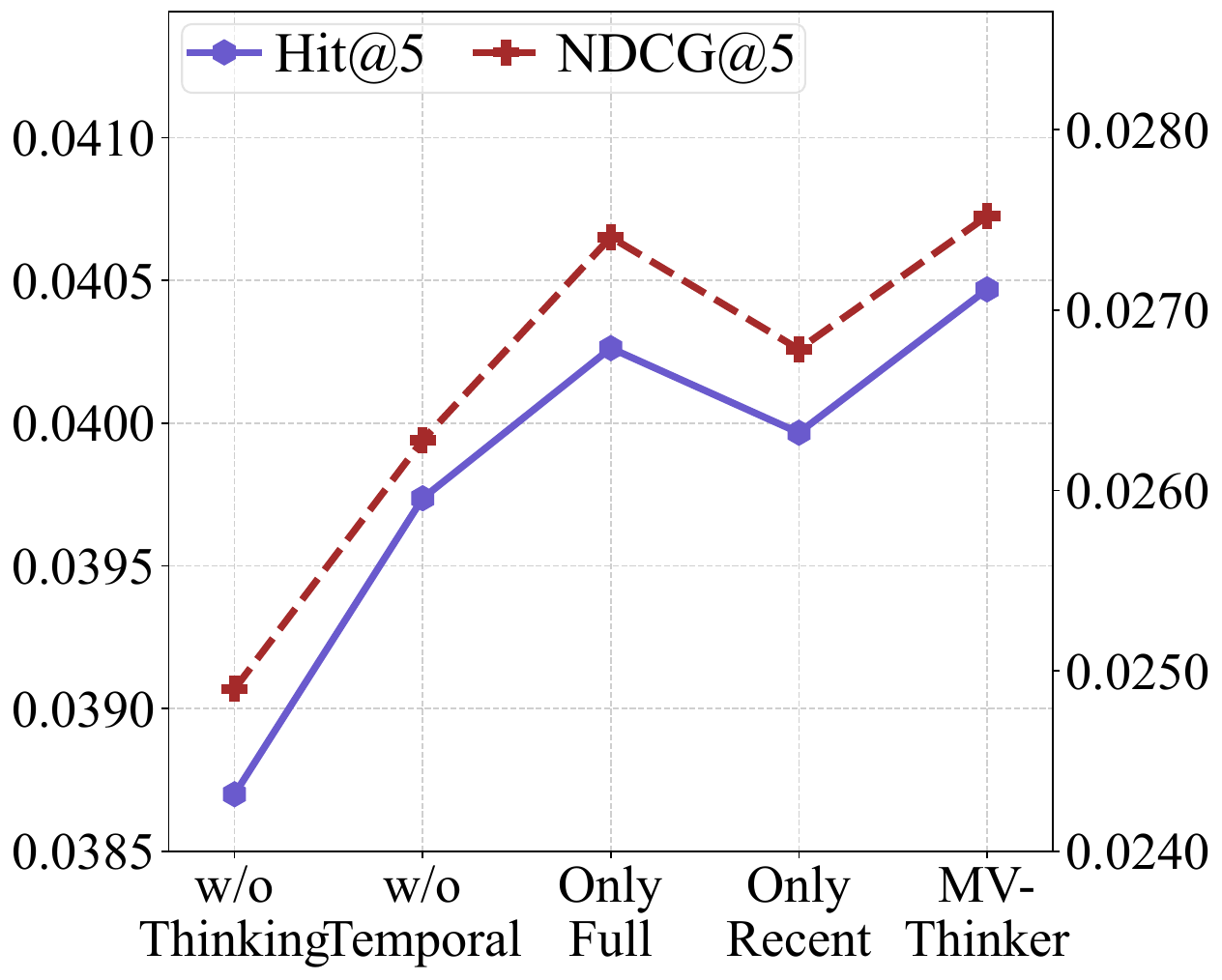}
\caption{Beauty}
\label{fig:component_beauty}
\end{subfigure}
\caption{Component analysis of \model{} on Instruments and Beauty.
``Only Full'' and ``Only Recent'' retain only the full-history
and recent-history views, respectively.}
\label{fig:component_analysis}
\vspace{-10pt}
\end{figure}

\subsection{Component Analysis (RQ5)}
\label{sec:component_analysis}

To investigate the effects of latent thinking, temporal modeling,
and the two history views, we compare w/o Thinking, w/o Temporal,
Only Full, and Only Recent on Instruments and Beauty.
The variants remove thinking or temporal modeling, or retain
only the full-history or recent-history view.
Figure~\ref{fig:component_analysis}(a--b) shows that \model{}
achieves the highest Hit@5 and NDCG@5 on both datasets.
Removing thinking causes the largest decline, and removing
temporal modeling also reduces performance. Only Full
outperforms Only Recent, but both trail the complete model.
These results suggest that latent thinking and temporal modeling
improve recommendation quality, with full-history evidence
providing a stronger foundation and recent-history cues adding
complementary information. \textbf{Combining both views through latent
thinking yields the best performance in these comparisons.}

\section{Related Work}
\label{sec:related}

\subsection{Generative Recommendation with Adaptive Generation}
\label{sec:related-gr}

Semantic-ID-based generative recommendation predicts next items through conditional identifier generation~\cite{wang2025generative,nawara2025comprehensive,deldjoo2024review,hou2026survey}, unifying retrieval and ranking~\cite{zhang2026tearag,han2026data,ji2024genrec,li2025ctr,li2024uncertaintyrag}. TIGER~\cite{rajput2023recommender} autoregressively generates content-derived semantic IDs. Subsequent work improves tokenization and generation~\cite{wang2024content,deng2025onerec,zheng2024adapting,xu2025mmq,xing2026reg4rec}: LETTER~\cite{wang2024learnable} integrates semantic and collaborative signals, SETRec~\cite{lin2025order} removes within-identifier token-order dependencies, and RPG~\cite{hou2025generating} generates long SIDs in parallel. General adaptation includes expert routing~\cite{shazeer2017outrageously,fedus2022switch,ma2018modeling,tang2020progressive}, dynamic depth~\cite{xin2020deebert,elbayad2019depth}, adapters and low-rank updates~\cite{houlsby2019parameter,hu2021lora}, module composition and conditional sharing~\cite{pfeiffer2021adapterfusion,mahabadi2021parameter,mao2022unipelt}, and continuous prompts~\cite{li2021prefix,lester2021power}. In generative retrieval, prefix pruning may misalign intermediate decisions with complete identifier quality~\cite{wu2025constrained}. APAO~\cite{yu2026apao} addresses this through adaptively weighted prefix objectives, while Latte~\cite{hou2026expressiveness} uses discrete latent prefixes to create multiple generation paths per item.

These studies improve identifier design, training objectives, and decoding flexibility. However, personalized contextual adaptation for progressive SID decisions remains underexplored. We adapt the SID-prediction context to the current user need rather than apply a fixed transformation across recommendation situations.

\subsection{Temporal Preference Modeling and Latent Thinking}
\label{sec:related-thinking}

Temporal preference modeling captures evolving interests through interaction order, time intervals, and historical context~\cite{sun2019bert4rec,li2020time,zhou2018deep,zhou2019deep,pi2020search}, while multi-interest models represent heterogeneous needs with multiple vectors~\cite{li2019multi,cen2020controllable}. SASRec~\cite{kang2018self} identifies relevant historical interactions through causal self-attention, whereas GEMs~\cite{zhou2026gems} organizes long histories into recent, mid-term, and lifecycle streams. Iterative and adaptive computation supports hidden-state refinement~\cite{dehghani2018universal,graves2016adaptive,banino2021pondernet}; language models explore continuous latent reasoning~\cite{hao2024training} and internal rationales before prediction~\cite{zelikman2024quiet}. ReaRec~\cite{tang2026think} and LARES~\cite{liu2025lares} iteratively refine hidden states, while CARE~\cite{lin2026bringing} employs query-anchored reasoning and progressively exposes finer-grained historical SID semantics before token generation.

These methods model temporal preferences within a sequence or separate streams, or refine the entire history. However, they rarely coordinate overlapping full- and recent-history evidence within a shared preference context for each user. We therefore use shared queries and view-specific readers to contextualize recent activity with the full history before SID generation.

\section{Conclusion}
\label{sec:conclusion}

In this paper, we propose \model{}, a semantic-ID-based generative recommendation framework that separates the organization of behavioral evidence from the adaptation of output-side transformations. Multi-View Temporal Thinking uses shared queries and independent readers to expose full-history and recent-history evidence before generation. Hierarchical Residual Adaptation supplies nested transformations of decoder states while retaining a common 
vocabulary projection, with loss-aware sampling during training and confidence-based selection at inference. 

\balance
\bibliographystyle{ACM-Reference-Format}
\bibliography{8.reference}

@inproceedings{ni-etal-2019-justifying,
    title = "Justifying Recommendations using Distantly-Labeled Reviews and Fine-Grained Aspects",
    author = "Ni, Jianmo  and
      Li, Jiacheng  and
      McAuley, Julian",
    editor = "Inui, Kentaro  and
      Jiang, Jing  and
      Ng, Vincent  and
      Wan, Xiaojun",
    booktitle = "Proceedings of the 2019 Conference on Empirical Methods in Natural Language Processing and the 9th International Joint Conference on Natural Language Processing (EMNLP-IJCNLP)",
    month = nov,
    year = "2019",
    address = "Hong Kong, China",
    publisher = "Association for Computational Linguistics",
    url = "https://aclanthology.org/D19-1018/",
    doi = "10.18653/v1/D19-1018",
    pages = "188--197"
}

@inproceedings{hou2025generating,
  title={Generating long semantic ids in parallel for recommendation},
  author={Hou, Yupeng and Li, Jiacheng and Shin, Ashley and Jeon, Jinsung and Santhanam, Abhishek and Shao, Wei and Hassani, Kaveh and Yao, Ning and McAuley, Julian},
  booktitle={Proceedings of the 31st ACM SIGKDD Conference on Knowledge Discovery and Data Mining V. 2},
  pages={956--966},
  year={2025}
}

@article{hidasi2015session,
  title={Session-based recommendations with recurrent neural networks},
  author={Hidasi, Bal{\'a}zs and Karatzoglou, Alexandros and Baltrunas, Linas and Tikk, Domonkos},
  journal={arXiv preprint arXiv:1511.06939},
  year={2015}
}

@article{li2025ctr,
  title={Ctr-sink: Attention sink for language models in click-through rate prediction},
  author={Li, Zixuan and Geng, Binzong and Xiong, Jing and He, Yong and Hu, Yuxuan and Chen, Jian and Chen, Dingwei and Chang, Xiyu and Wong, Ngai and Zhang, Liang and others},
  journal={arXiv preprint arXiv:2508.03668},
  year={2025}
}

@inproceedings{wang2025generative,
  title={Generative large recommendation models: Emerging trends in llms for recommendation},
  author={Wang, Hao and Guo, Wei and Zhang, Luankang and Chin, Jin Yao and Ye, Yufei and Guo, Huifeng and Liu, Yong and Lian, Defu and Tang, Ruiming and Chen, Enhong},
  booktitle={Companion Proceedings of the ACM on Web Conference 2025},
  pages={49--52},
  year={2025}
}

@article{nawara2025comprehensive,
  title={A comprehensive survey on LLM-powered recommender systems: From discriminative, generative to multi-modal paradigms},
  author={Nawara, Dina and Kashef, Rasha},
  journal={IEEE Access},
  year={2025},
  publisher={IEEE}
}

@inproceedings{ji2024genrec,
  title={Genrec: Large language model for generative recommendation},
  author={Ji, Jianchao and Li, Zelong and Xu, Shuyuan and Hua, Wenyue and Ge, Yingqiang and Tan, Juntao and Zhang, Yongfeng},
  booktitle={European Conference on Information Retrieval},
  pages={494--502},
  year={2024},
  organization={Springer}
}

@inproceedings{deldjoo2024review,
  title={A review of modern recommender systems using generative models (gen-recsys)},
  author={Deldjoo, Yashar and He, Zhankui and McAuley, Julian and Korikov, Anton and Sanner, Scott and Ramisa, Arnau and Vidal, Ren{\'e} and Sathiamoorthy, Maheswaran and Kasirzadeh, Atoosa and Milano, Silvia},
  booktitle={Proceedings of the 30th ACM SIGKDD conference on Knowledge Discovery and Data Mining},
  pages={6448--6458},
  year={2024}
}

@article{hou2026survey,
  title={A survey on generative recommendation: Data, model, and tasks},
  author={Hou, Min and Wu, Le and Liao, Yuxin and Yang, Yonghui and Zhang, Zhen and Wang, Yu and Zheng, Changlong and Wu, Han and Hong, Richang},
  journal={AI Open},
  year={2026},
  publisher={Elsevier}
}

@article{lipowski2012roulette,
  title={Roulette-wheel selection via stochastic acceptance},
  author={Lipowski, Adam and Lipowska, Dorota},
  journal={Physica A: Statistical Mechanics and its Applications},
  volume={391},
  number={6},
  pages={2193--2196},
  year={2012},
  publisher={Elsevier}
}

@inproceedings{wang2024content,
  title={Content-based collaborative generation for recommender systems},
  author={Wang, Yidan and Ren, Zhaochun and Sun, Weiwei and Yang, Jiyuan and Liang, Zhixiang and Chen, Xin and Xie, Ruobing and Yan, Su and Zhang, Xu and Ren, Pengjie and others},
  booktitle={Proceedings of the 33rd ACM International Conference on Information and Knowledge Management},
  pages={2420--2430},
  year={2024}
}

@article{deng2025onerec,
  title={Onerec: Unifying retrieve and rank with generative recommender and iterative preference alignment},
  author={Deng, Jiaxin and Wang, Shiyao and Cai, Kuo and Ren, Lejian and Hu, Qigen and Ding, Weifeng and Luo, Qiang and Zhou, Guorui},
  journal={arXiv preprint arXiv:2502.18965},
  year={2025}
}

@inproceedings{zheng2024adapting,
  title={Adapting large language models by integrating collaborative semantics for recommendation},
  author={Zheng, Bowen and Hou, Yupeng and Lu, Hongyu and Chen, Yu and Zhao, Wayne Xin and Chen, Ming and Wen, Ji-Rong},
  booktitle={2024 IEEE 40th International Conference on Data Engineering (ICDE)},
  pages={1435--1448},
  year={2024},
  organization={IEEE}
}

@article{xu2025mmq,
  title={Mmq-v2: Align, denoise, and amplify: Adaptive behavior mining for semantic ids learning in recommendation},
  author={Xu, Yi and Zhang, Moyu and Fan, Chaofan and Hu, Jinxin and Li, Xiaochen and Zhang, Yu and Zeng, Xiaoyi and Zhang, Jing},
  journal={arXiv e-prints},
  pages={arXiv--2510},
  year={2025}
}

@inproceedings{xing2026reg4rec,
  title={Reg4rec: Reasoning-enhanced generative model for large-scale recommendation systems},
  author={Xing, Haibo and Deng, Hao and Mao, Yucheng and Mu, Lingyu and Hu, Jinxin and Xu, Yi and Zhang, Hao and Wang, Jiahao and Wang, Shizhun and Zhang, Yu and others},
  booktitle={2026 IEEE 42nd International Conference on Data Engineering (ICDE)},
  pages={3799--3812},
  year={2026},
  organization={IEEE}
}

@inproceedings{lin2025order,
  title={Order-agnostic identifier for large language model-based generative recommendation},
  author={Lin, Xinyu and Shi, Haihan and Wang, Wenjie and Feng, Fuli and Wang, Qifan and Ng, See-Kiong and Chua, Tat-Seng},
  booktitle={Proceedings of the 48th international ACM SIGIR conference on research and development in information retrieval},
  pages={1923--1933},
  year={2025}
}

@article{yang2026reasoning,
  title={Where Reasoning Matters: Rethinking Latent Reasoning in Semantic ID-based Generative Recommendation},
  author={Yang, Shangxin and Gao, Min and Wang, Zongwei and Yu, Junliang},
  journal={arXiv preprint arXiv:2607.12425},
  year={2026}
}

@inproceedings{kang2018self,
  title={Self-attentive sequential recommendation},
  author={Kang, Wang-Cheng and McAuley, Julian},
  booktitle={2018 IEEE international conference on data mining (ICDM)},
  pages={197--206},
  year={2018},
  organization={IEEE}
}

@article{li2024uncertaintyrag,
  title={Uncertaintyrag: Span-level uncertainty enhanced long-context modeling for retrieval-augmented generation},
  author={Li, Zixuan and Xiong, Jing and Ye, Fanghua and Zheng, Chuanyang and Wu, Xun and Lu, Jianqiao and Wan, Zhongwei and Liang, Xiaodan and Li, Chengming and Sun, Zhenan and others},
  journal={arXiv preprint arXiv:2410.02719},
  year={2024}
}

@article{zhang2026tearag,
  title={Tearag: A token-efficient agentic retrieval-augmented generation framework},
  author={Zhang, Chao and Wang, Yuhao and Xu, Derong and Zhang, Haoxin and Lyu, Yuanjie and Chen, Yuhao and Liu, Shuochen and Xu, Tong and Zhao, Xiangyu and Gao, Yan and others},
  journal={ACM Transactions on Information Systems},
  volume={44},
  number={6},
  pages={1--35},
  year={2026},
  publisher={ACM New York, NY}
}

@article{han2026data,
  title={Data efficient adaptation in large language models via continuous low-rank fine-tuning},
  author={Han, Xiao and Zhao, Zimo and Wang, Wanyu and Wang, Maolin and Liu, Zitao and Chang, Yi and Zhao, Xiangyu},
  journal={Advances in Neural Information Processing Systems},
  volume={38},
  pages={165157--165182},
  year={2026}
}

@inproceedings{sun2019bert4rec,
  title={BERT4Rec: Sequential recommendation with bidirectional encoder representations from transformer},
  author={Sun, Fei and Liu, Jun and Wu, Jian and Pei, Changhua and Lin, Xiao and Ou, Wenwu and Jiang, Peng},
  booktitle={Proceedings of the 28th ACM international conference on information and knowledge management},
  pages={1441--1450},
  year={2019}
}

@article{tang2026think,
  title={Think before recommend: Unleashing the latent reasoning power for sequential recommendation},
  author={Tang, Jiakai and Dai, Sunhao and Shi, Teng and Xu, Jun and Chen, Xu and Chen, Wen and Wu, Jian and Jiang, Yuning},
  journal={IEEE Transactions on Knowledge and Data Engineering},
  year={2026},
  publisher={IEEE}
}

@article{liu2025lares,
  title={Lares: Latent reasoning for sequential recommendation},
  author={Liu, Enze and Zheng, Bowen and Wang, Xiaolei and Zhao, Wayne Xin and Wang, Jinpeng and Chen, Sheng and Wen, Ji-Rong},
  journal={arXiv preprint arXiv:2505.16865},
  year={2025}
}

@inproceedings{wu2025constrained,
  title={Constrained auto-regressive decoding constrains generative retrieval},
  author={Wu, Shiguang and Ren, Zhaochun and Xin, Xin and Yang, Jiyuan and Zhang, Mengqi and Chen, Zhumin and de Rijke, Maarten and Ren, Pengjie},
  booktitle={Proceedings of the 48th International ACM SIGIR Conference on Research and Development in Information Retrieval},
  pages={2429--2440},
  year={2025}
}

@article{rajput2023recommender,
  title={Recommender systems with generative retrieval},
  author={Rajput, Shashank and Mehta, Nikhil and Singh, Anima and Hulikal Keshavan, Raghunandan and Vu, Trung and Heldt, Lukasz and Hong, Lichan and Tay, Yi and Tran, Vinh and Samost, Jonah and others},
  journal={Advances in Neural Information Processing Systems},
  volume={36},
  pages={10299--10315},
  year={2023}
}

@inproceedings{wang2024learnable,
  title={Learnable item tokenization for generative recommendation},
  author={Wang, Wenjie and Bao, Honghui and Lin, Xinyu and Zhang, Jizhi and Li, Yongqi and Feng, Fuli and Ng, See-Kiong and Chua, Tat-Seng},
  booktitle={Proceedings of the 33rd ACM International Conference on Information and Knowledge Management},
  pages={2400--2409},
  year={2024}
}

@inproceedings{lin2026bringing,
  title={Bringing reasoning to generative recommendation through the lens of cascaded ranking},
  author={Lin, Xinyu and Liu, Pengyuan and Wang, Wenjie and Hu, Yicheng and Xu, Chen and Feng, Fuli and Wang, Qifan and Chua, Tat-Seng},
  booktitle={Proceedings of the ACM Web Conference 2026},
  pages={8939--8949},
  year={2026}
}

@article{zhou2026gems,
  title={Gems: Breaking the long-sequence barrier in generative recommendation with a multi-stream decoder},
  author={Zhou, Yu and Guo, Chengcheng and Cai, Kuo and Liu, Ji and Luo, Qiang and Tang, Ruiming and Li, Han and Gai, Kun and Zhou, Guorui},
  journal={arXiv preprint arXiv:2602.13631},
  year={2026}
}

@inproceedings{yu2026apao,
  title={APAO: Bridging the Training-Inference Gap in Generative Recommendation via Adaptive Prefix-Aware Optimization},
  author={Yu, Yuanqing and Wang, Yifan and Ma, Weizhi and Guo, Zhiqiang and Zhang, Min},
  booktitle={Proceedings of the 32nd ACM SIGKDD Conference on Knowledge Discovery and Data Mining V. 2},
  pages={6338--6349},
  year={2026}
}

@article{hou2026expressiveness,
  title={Expressiveness Limits of Autoregressive Semantic ID Generation in Generative Recommendation},
  author={Hou, Yupeng and Kim, Haven and Ju, Clark Mingxuan and Escoto, Eduardo and Shah, Neil and McAuley, Julian},
  journal={arXiv preprint arXiv:2605.06331},
  year={2026}
}

@inproceedings{tang2018personalized,
  title={Personalized top-n sequential recommendation via convolutional sequence embedding},
  author={Tang, Jiaxi and Wang, Ke},
  booktitle={Proceedings of the eleventh ACM international conference on web search and data mining},
  pages={565--573},
  year={2018}
}

@inproceedings{li2020time,
  title={Time interval aware self-attention for sequential recommendation},
  author={Li, Jiacheng and Wang, Yujie and McAuley, Julian},
  booktitle={Proceedings of the 13th international conference on web search and data mining},
  pages={322--330},
  year={2020}
}

@inproceedings{zhou2018deep,
  title={Deep interest network for click-through rate prediction},
  author={Zhou, Guorui and Zhu, Xiaoqiang and Song, Chenru and Fan, Ying and Zhu, Han and Ma, Xiao and Yan, Yanghui and Jin, Junqi and Li, Han and Gai, Kun},
  booktitle={Proceedings of the 24th ACM SIGKDD international conference on knowledge discovery \& data mining},
  pages={1059--1068},
  year={2018}
}

@inproceedings{zhou2019deep,
  title={Deep interest evolution network for click-through rate prediction},
  author={Zhou, Guorui and Mou, Na and Fan, Ying and Pi, Qi and Bian, Weijie and Zhou, Chang and Zhu, Xiaoqiang and Gai, Kun},
  booktitle={Proceedings of the AAAI conference on artificial intelligence},
  volume={33},
  number={01},
  pages={5941--5948},
  year={2019}
}

@inproceedings{li2019multi,
  title={Multi-interest network with dynamic routing for recommendation at Tmall},
  author={Li, Chao and Liu, Zhiyuan and Wu, Mengmeng and Xu, Yuchi and Zhao, Huan and Huang, Pipei and Kang, Guoliang and Chen, Qiwei and Li, Wei and Lee, Dik Lun},
  booktitle={Proceedings of the 28th ACM international conference on information and knowledge management},
  pages={2615--2623},
  year={2019}
}

@inproceedings{cen2020controllable,
  title={Controllable multi-interest framework for recommendation},
  author={Cen, Yukuo and Zhang, Jianwei and Zou, Xu and Zhou, Chang and Yang, Hongxia and Tang, Jie},
  booktitle={Proceedings of the 26th ACM SIGKDD international conference on knowledge discovery \& data mining},
  pages={2942--2951},
  year={2020}
}

@inproceedings{pi2020search,
  title={Search-based user interest modeling with lifelong sequential behavior data for click-through rate prediction},
  author={Pi, Qi and Zhou, Guorui and Zhang, Yujing and Wang, Zhe and Ren, Lejian and Fan, Ying and Zhu, Xiaoqiang and Gai, Kun},
  booktitle={Proceedings of the 29th ACM International Conference on Information \& Knowledge Management},
  pages={2685--2692},
  year={2020}
}

@article{dehghani2018universal,
  title={Universal transformers},
  author={Dehghani, Mostafa and Gouws, Stephan and Vinyals, Oriol and Uszkoreit, Jakob and Kaiser, {\L}ukasz},
  journal={arXiv preprint arXiv:1807.03819},
  year={2018}
}

@article{graves2016adaptive,
  title={Adaptive computation time for recurrent neural networks},
  author={Graves, Alex},
  journal={arXiv preprint arXiv:1603.08983},
  year={2016}
}

@article{banino2021pondernet,
  title={Pondernet: Learning to ponder},
  author={Banino, Andrea and Balaguer, Jan and Blundell, Charles},
  journal={arXiv preprint arXiv:2107.05407},
  year={2021}
}

@article{hao2024training,
  title={Training large language models to reason in a continuous latent space},
  author={Hao, Shibo and Sukhbaatar, Sainbayar and Su, DiJia and Li, Xian and Hu, Zhiting and Weston, Jason and Tian, Yuandong},
  journal={arXiv preprint arXiv:2412.06769},
  year={2024}
}

@article{zelikman2024quiet,
  title={Quiet-star: Language models can teach themselves to think before speaking},
  author={Zelikman, Eric and Harik, Georges and Shao, Yijia and Jayasiri, Varuna and Haber, Nick and Goodman, Noah D},
  journal={arXiv preprint arXiv:2403.09629},
  year={2024}
}

@inproceedings{houlsby2019parameter,
  title={Parameter-efficient transfer learning for NLP},
  author={Houlsby, Neil and Giurgiu, Andrei and Jastrzebski, Stanislaw and Morrone, Bruna and De Laroussilhe, Quentin and Gesmundo, Andrea and Attariyan, Mona and Gelly, Sylvain},
  booktitle={International conference on machine learning},
  pages={2790--2799},
  year={2019},
  organization={PMLR}
}

@article{hu2021lora,
  title={Lora: Low-rank adaptation of large language models},
  author={Hu, Edward J and Shen, Yelong and Wallis, Phillip and Allen-Zhu, Zeyuan and Li, Yuanzhi and Wang, Shean and Wang, Lu and Chen, Weizhu},
  journal={arXiv preprint arXiv:2106.09685},
  year={2021}
}

@inproceedings{pfeiffer2021adapterfusion,
  title={Adapterfusion: Non-destructive task composition for transfer learning},
  author={Pfeiffer, Jonas and Kamath, Aishwarya and R{\"u}ckl{\'e}, Andreas and Cho, Kyunghyun and Gurevych, Iryna},
  booktitle={Proceedings of the 16th conference of the European chapter of the association for computational linguistics: main volume},
  pages={487--503},
  year={2021}
}

@inproceedings{mahabadi2021parameter,
  title={Parameter-efficient multi-task fine-tuning for transformers via shared hypernetworks},
  author={Mahabadi, Rabeeh Karimi and Ruder, Sebastian and Dehghani, Mostafa and Henderson, James},
  booktitle={Proceedings of the 59th Annual Meeting of the Association for Computational Linguistics and the 11th International Joint Conference on Natural Language Processing (Volume 1: Long Papers)},
  pages={565--576},
  year={2021}
}

@inproceedings{li2021prefix,
  title={Prefix-tuning: Optimizing continuous prompts for generation},
  author={Li, Xiang Lisa and Liang, Percy},
  booktitle={Proceedings of the 59th annual meeting of the association for computational linguistics and the 11th international joint conference on natural language processing (volume 1: Long papers)},
  pages={4582--4597},
  year={2021}
}

@inproceedings{lester2021power,
  title={The power of scale for parameter-efficient prompt tuning},
  author={Lester, Brian and Al-Rfou, Rami and Constant, Noah},
  booktitle={Proceedings of the 2021 conference on empirical methods in natural language processing},
  pages={3045--3059},
  year={2021}
}

@inproceedings{mao2022unipelt,
  title={Unipelt: A unified framework for parameter-efficient language model tuning},
  author={Mao, Yuning and Mathias, Lambert and Hou, Rui and Almahairi, Amjad and Ma, Hao and Han, Jiawei and Yih, Wen-tau and Khabsa, Madian},
  booktitle={Proceedings of the 60th Annual Meeting of the Association for Computational Linguistics (Volume 1: Long Papers)},
  pages={6253--6264},
  year={2022}
}

@inproceedings{ma2018modeling,
  title={Modeling task relationships in multi-task learning with multi-gate mixture-of-experts},
  author={Ma, Jiaqi and Zhao, Zhe and Yi, Xinyang and Chen, Jilin and Hong, Lichan and Chi, Ed H},
  booktitle={Proceedings of the 24th ACM SIGKDD international conference on knowledge discovery \& data mining},
  pages={1930--1939},
  year={2018}
}

@inproceedings{tang2020progressive,
  title={Progressive layered extraction (ple): A novel multi-task learning (mtl) model for personalized recommendations},
  author={Tang, Hongyan and Liu, Junning and Zhao, Ming and Gong, Xudong},
  booktitle={Proceedings of the 14th ACM conference on recommender systems},
  pages={269--278},
  year={2020}
}

@article{shazeer2017outrageously,
  title={Outrageously large neural networks: The sparsely-gated mixture-of-experts layer},
  author={Shazeer, Noam and Mirhoseini, Azalia and Maziarz, Krzysztof and Davis, Andy and Le, Quoc and Hinton, Geoffrey and Dean, Jeff},
  journal={arXiv preprint arXiv:1701.06538},
  year={2017}
}

@article{fedus2022switch,
  title={Switch transformers: Scaling to trillion parameter models with simple and efficient sparsity},
  author={Fedus, William and Zoph, Barret and Shazeer, Noam},
  journal={Journal of Machine Learning Research},
  volume={23},
  number={120},
  pages={1--39},
  year={2022}
}

@article{elbayad2019depth,
  title={Depth-adaptive transformer},
  author={Elbayad, Maha and Gu, Jiatao and Grave, Edouard and Auli, Michael},
  journal={arXiv preprint arXiv:1910.10073},
  year={2019}
}

@inproceedings{xin2020deebert,
  title={DeeBERT: Dynamic early exiting for accelerating BERT inference},
  author={Xin, Ji and Tang, Raphael and Lee, Jaejun and Yu, Yaoliang and Lin, Jimmy},
  booktitle={Proceedings of the 58th annual meeting of the association for computational linguistics},
  pages={2246--2251},
  year={2020}
}

@article{vaswani2017attention,
  title={Attention is all you need},
  author={Vaswani, Ashish and Shazeer, Noam and Parmar, Niki and Uszkoreit, Jakob and Jones, Llion and Gomez, Aidan N and Kaiser, {\L}ukasz and Polosukhin, Illia},
  journal={Advances in neural information processing systems},
  volume={30},
  year={2017}
}

@article{ba2016layer,
  title={Layer normalization},
  author={Ba, Jimmy Lei and Kiros, Jamie Ryan and Hinton, Geoffrey E},
  journal={arXiv preprint arXiv:1607.06450},
  year={2016}
}

@article{vijayakumar2016diverse,
  title={Diverse beam search: Decoding diverse solutions from neural sequence models},
  author={Vijayakumar, Ashwin K and Cogswell, Michael and Selvaraju, Ramprasath R and Sun, Qing and Lee, Stefan and Crandall, David and Batra, Dhruv},
  journal={arXiv preprint arXiv:1610.02424},
  year={2016}
}

@inproceedings{covington2016deep,
  title={Deep neural networks for youtube recommendations},
  author={Covington, Paul and Adams, Jay and Sargin, Emre},
  booktitle={Proceedings of the 10th ACM conference on recommender systems},
  pages={191--198},
  year={2016}
}

@inproceedings{zhao2019recommending,
  title={Recommending what video to watch next: a multitask ranking system},
  author={Zhao, Zhe and Hong, Lichan and Wei, Li and Chen, Jilin and Nath, Aniruddh and Andrews, Shawn and Kumthekar, Aditee and Sathiamoorthy, Maheswaran and Yi, Xinyang and Chi, Ed},
  booktitle={Proceedings of the 13th ACM conference on recommender systems},
  pages={43--51},
  year={2019}
}

@inproceedings{geng2022recommendation,
  title={Recommendation as language processing (rlp): A unified pretrain, personalized prompt \& predict paradigm (p5)},
  author={Geng, Shijie and Liu, Shuchang and Fu, Zuohui and Ge, Yingqiang and Zhang, Yongfeng},
  booktitle={Proceedings of the 16th ACM conference on recommender systems},
  pages={299--315},
  year={2022}
}

@inproceedings{liao2024llara,
  title={Llara: Large language-recommendation assistant},
  author={Liao, Jiayi and Li, Sihang and Yang, Zhengyi and Wu, Jiancan and Yuan, Yancheng and Wang, Xiang and He, Xiangnan},
  booktitle={Proceedings of the 47th international ACM SIGIR conference on research and development in information retrieval},
  pages={1785--1795},
  year={2024}
}

@article{bao2025bi,
  title={A bi-step grounding paradigm for large language models in recommendation systems},
  author={Bao, Keqin and Zhang, Jizhi and Wang, Wenjie and Zhang, Yang and Yang, Zhengyi and Luo, Yanchen and Chen, Chong and Feng, Fuli and Tian, Qi},
  journal={ACM Transactions on Recommender Systems},
  volume={3},
  number={4},
  pages={1--27},
  year={2025},
  publisher={ACM New York, NY}
}

@inproceedings{wang2024eager,
  title={Eager: Two-stream generative recommender with behavior-semantic collaboration},
  author={Wang, Ye and Xun, Jiahao and Hong, Minjie and Zhu, Jieming and Jin, Tao and Lin, Wang and Li, Haoyuan and Li, Linjun and Xia, Yan and Zhao, Zhou and others},
  booktitle={Proceedings of the 30th ACM SIGKDD Conference on Knowledge Discovery and Data Mining},
  pages={3245--3254},
  year={2024}
}

@article{villatel2018recurrent,
  title={Recurrent neural networks for long and short-term sequential recommendation},
  author={Villatel, Kiewan and Smirnova, Elena and Mary, J{\'e}r{\'e}mie and Preux, Philippe},
  journal={arXiv preprint arXiv:1807.09142},
  year={2018}
}

@inproceedings{li2017neural,
  title={Neural attentive session-based recommendation},
  author={Li, Jing and Ren, Pengjie and Chen, Zhumin and Ren, Zhaochun and Lian, Tao and Ma, Jun},
  booktitle={Proceedings of the 2017 ACM on Conference on Information and Knowledge Management},
  pages={1419--1428},
  year={2017}
}

@inproceedings{liu2018stamp,
  title={STAMP: short-term attention/memory priority model for session-based recommendation},
  author={Liu, Qiao and Zeng, Yifu and Mokhosi, Refuoe and Zhang, Haibin},
  booktitle={Proceedings of the 24th ACM SIGKDD international conference on knowledge discovery \& data mining},
  pages={1831--1839},
  year={2018}
}

@inproceedings{sun2020go,
  title={Where to go next: Modeling long-and short-term user preferences for point-of-interest recommendation},
  author={Sun, Ke and Qian, Tieyun and Chen, Tong and Liang, Yile and Nguyen, Quoc Viet Hung and Yin, Hongzhi},
  booktitle={Proceedings of the AAAI conference on artificial intelligence},
  volume={34},
  number={01},
  pages={214--221},
  year={2020}
}

@inproceedings{tan2021sparse,
  title={Sparse-interest network for sequential recommendation},
  author={Tan, Qiaoyu and Zhang, Jianwei and Yao, Jiangchao and Liu, Ninghao and Zhou, Jingren and Yang, Hongxia and Hu, Xia},
  booktitle={Proceedings of the 14th ACM international conference on web search and data mining},
  pages={598--606},
  year={2021}
}

@inproceedings{chang2021sequential,
  title={Sequential recommendation with graph neural networks},
  author={Chang, Jianxin and Gao, Chen and Zheng, Yu and Hui, Yiqun and Niu, Yanan and Song, Yang and Jin, Depeng and Li, Yong},
  booktitle={Proceedings of the 44th international ACM SIGIR conference on research and development in information retrieval},
  pages={378--387},
  year={2021}
}

@article{raffel2020exploring,
  title={Exploring the limits of transfer learning with a unified text-to-text transformer},
  author={Raffel, Colin and Shazeer, Noam and Roberts, Adam and Lee, Katherine and Narang, Sharan and Matena, Michael and Zhou, Yanqi and Li, Wei and Liu, Peter J},
  journal={Journal of machine learning research},
  volume={21},
  number={140},
  pages={1--67},
  year={2020}
}

@article{loshchilov2017decoupled,
  title={Decoupled weight decay regularization},
  author={Loshchilov, Ilya},
  journal={arXiv preprint arXiv:1711.05101},
  year={2017}
}

\appendix
\section{Training and Inference Algorithm}
\label{app:algorithm}

Algorithm~\ref{alg:mvthinker} summarizes the progressive training and confidence-based inference of \model{} described in Section~\ref{sec:training_inference}.

\begin{algorithm}[h]
\setlength\abovecaptionskip{0.2\baselineskip}
\setlength\belowcaptionskip{0.2\baselineskip}
\caption{Training and Inference Process of \model{}}
\label{alg:mvthinker}
\begin{algorithmic}[1]
\State \textbf{Input:} training set $\mathcal{D}$ and tokenized item SIDs
\State \textbf{Initialize:} encoder--decoder backbone
\Statex \textbf{Training Process}
\State \textbf{Stage 0:} Train the backbone with $\mathcal{L}_{\mathrm{GR}}$
\State \textbf{Stage 1:} Add thinking tokens and nested residual paths
\For{each batch $\mathcal{B}\subseteq\mathcal{D}$}
    \State Compute the sequence loss of each path
    \State Sample a path for each example by Eq.~\eqref{eq:roulette}
    \State Update the backbone, queries, and adapters by
    Eq.~\eqref{eq:hra_loss}
\EndFor
\State \textbf{Stage 2:} Add temporal readers and freeze the backbone
\For{each batch $\mathcal{B}\subseteq\mathcal{D}$}
    \State Construct two views and enhance their queries
    \State Compute the sequence loss of each path
    \State Sample a path for each example by Eq.~\eqref{eq:roulette}
    \State Update the readers, queries, and adapters by
    Eq.~\eqref{eq:hra_loss}
\EndFor
\Statex \textbf{Inference Process}
\For{each user $u$}
    \State Encode $H_u$ and enhance the thinking queries
    \State Select $k_u^*$ by Eq.~\eqref{eq:confidence-routing}
    \State Run constrained beam search with $k_u^*$ fixed
    \State Decode valid SIDs and return the ranked items
\EndFor
\end{algorithmic}
\end{algorithm}

\section{Catalog Coverage and Repeated Exposure}
\label{sec:coverage-exposure}

\begin{figure}[t]
\setlength\abovecaptionskip{0.2\baselineskip}
\setlength\belowcaptionskip{0.2\baselineskip}
\centering
\begin{minipage}[t]{0.49\linewidth}
\centering
\includegraphics[width=\linewidth]
{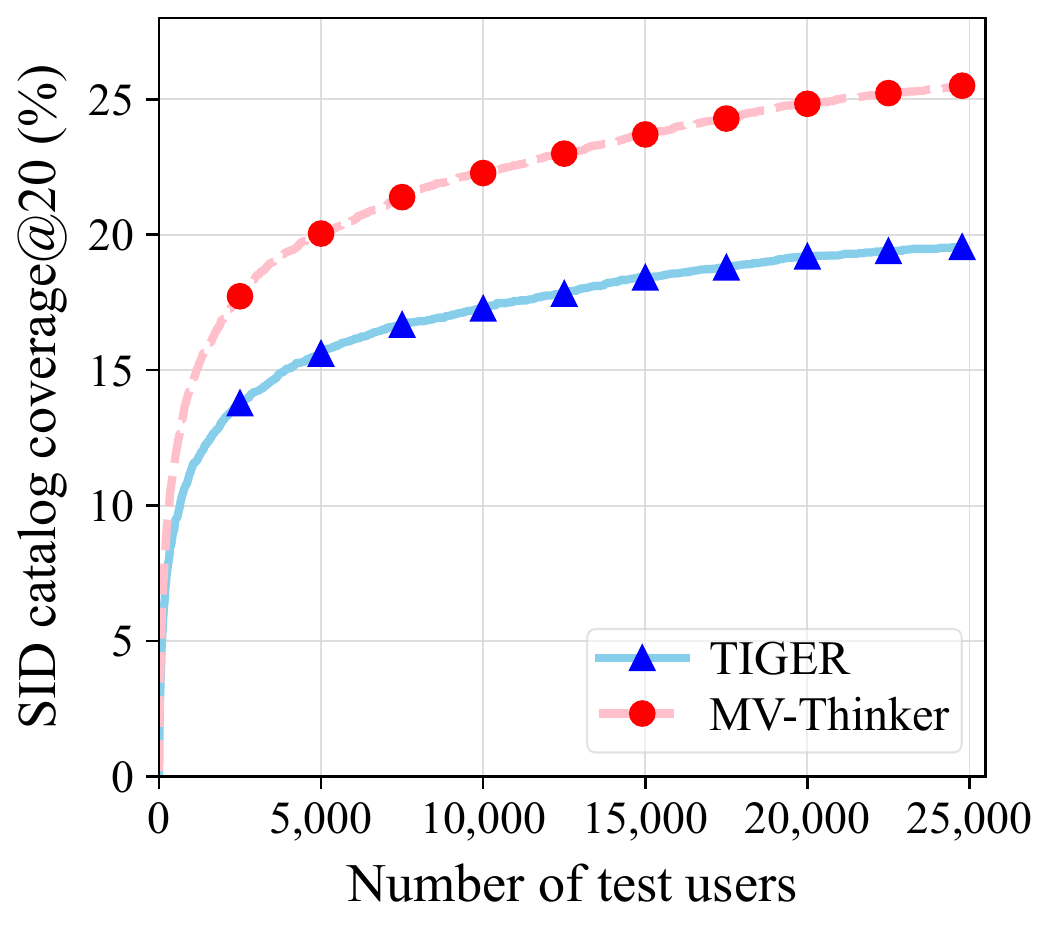}
\small (a) Cumulative SID coverage
\end{minipage}\hfill%
\begin{minipage}[t]{0.49\linewidth}
\centering
\includegraphics[width=\linewidth]
{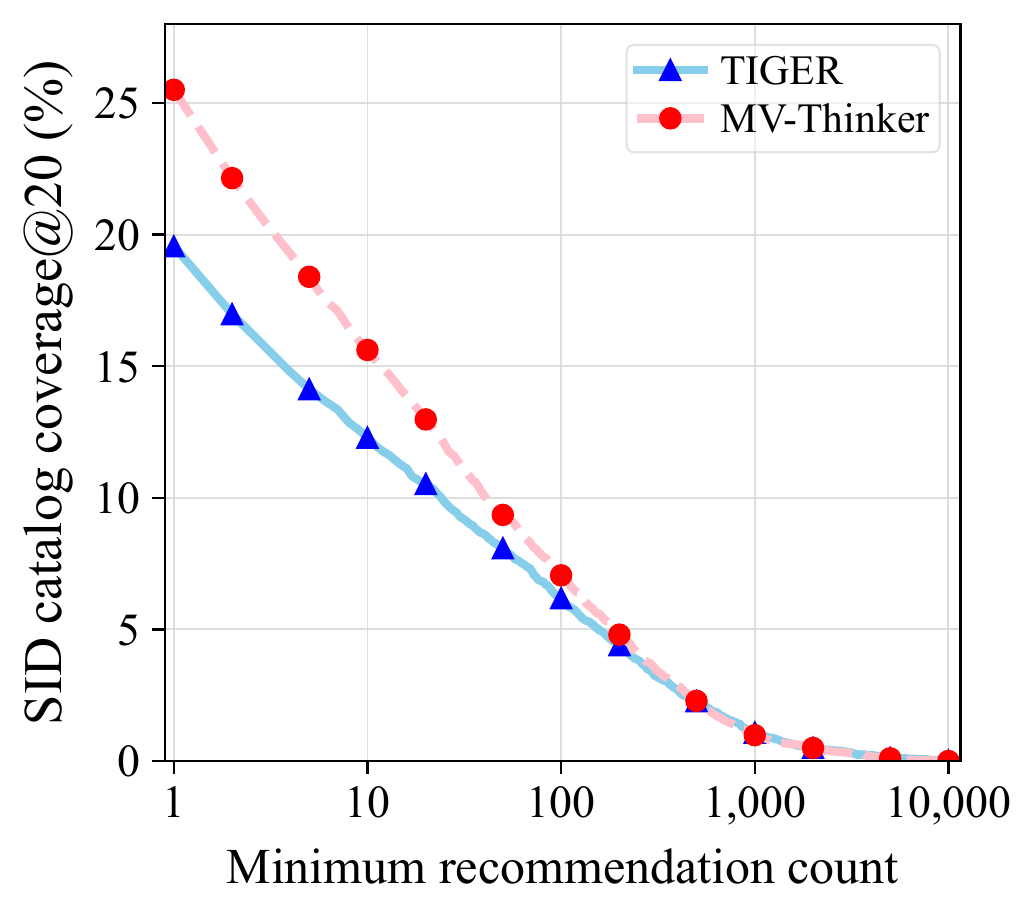}
\small (b) Frequency-threshold coverage
\end{minipage}
\caption{SID coverage and repeated recommendation exposure on
Instruments dataset at $K=20$. (a) Cumulative SID coverage in the same fixed
random user order. (b) Coverage of SIDs recommended to at least the
specified number of users (logarithmic horizontal axis).
}
\label{fig:recommendation-coverage}
\end{figure}

To examine whether broader catalog coverage extends beyond one-off
recommendations, we compare \model{} and TIGER on Instruments
test users using the same RQ-VAE index, histories, and constrained
beam search (width 20), returning 20 distinct valid SIDs per user.
Figure~\ref{fig:recommendation-coverage}(a) shows higher cumulative
coverage for \model{}, reaching 25.50\% versus 19.56\%.
Panel~(b) further shows that 1,545 SIDs reach at least 10 users,
compared with 1,216 for TIGER, although the advantage is not uniform
at high thresholds. Together with the ranking gains in
Table~\ref{tab:overall_performance}, these results show that
\model{} extends repeated exposure to more SIDs without sacrificing
overall recommendation accuracy, rather than merely increasing
one-off outputs. They suggest that coordinating full-history and
recent-history evidence helps provide a more informative shared
context for SID generation. By preserving recent cues alongside
the historical background needed to interpret them, this design
supports recommendations that reflect current interests without
discarding useful longer-term preferences.

\section{Hyper-Parameter Analysis}
\label{sec:parameter_analysis}
\begin{figure}[t]
\setlength\abovecaptionskip{0.2\baselineskip}
\setlength\belowcaptionskip{0.2\baselineskip}
\centering
\includegraphics[width=\columnwidth]{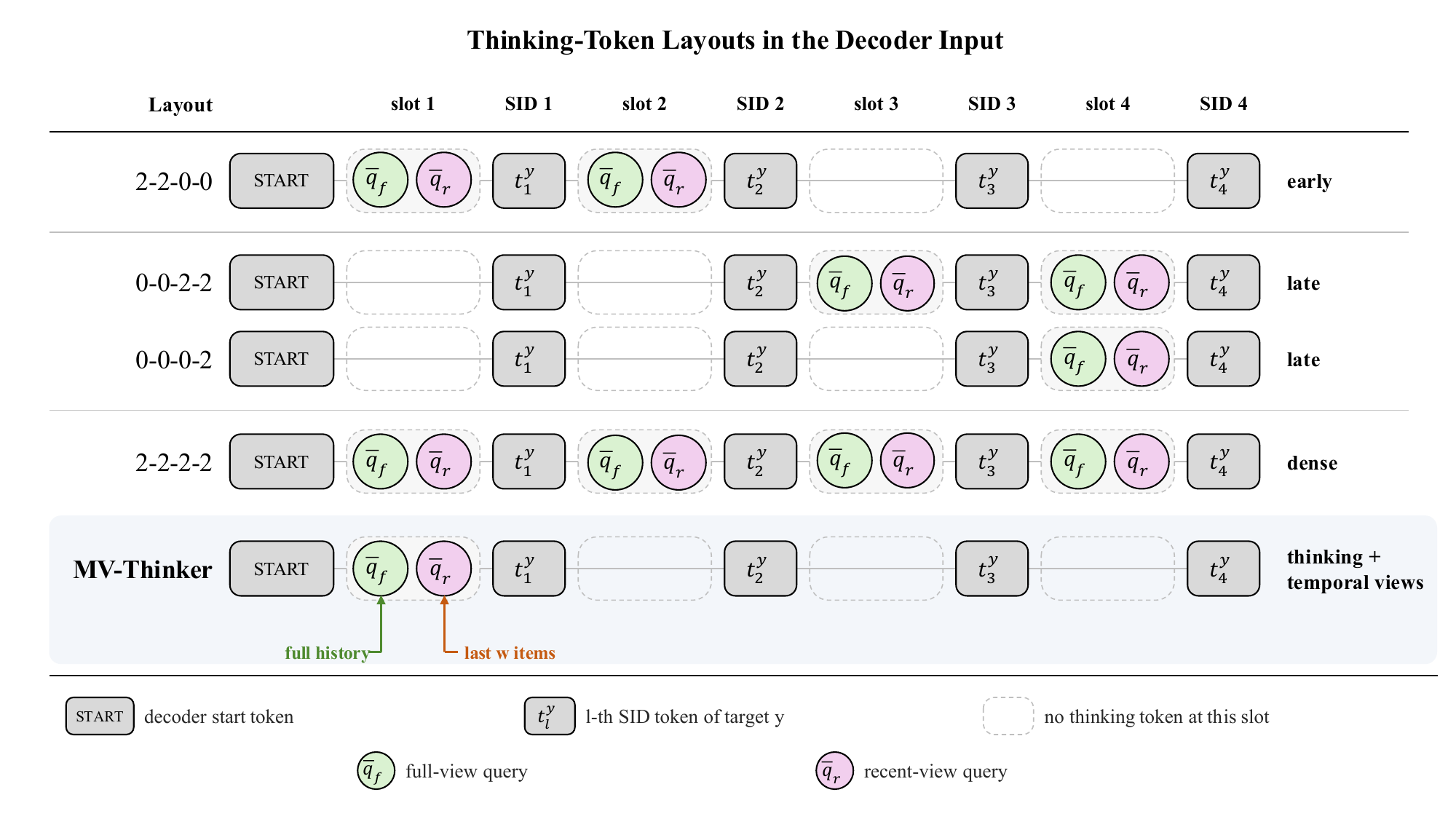}
\caption{Thinking-token layouts compared in Table~\ref{tab:thinking_layouts}. In $a$-$b$-$c$-$d$, each number denotes the thinking tokens inserted before the corresponding target SID token. Only \model{} reads the full-history and recent-history views.}
\label{fig:thinking_layouts}
\end{figure}

To investigate how thinking placement affects recommendation
performance, we compare \model{} with four positional layouts, as
illustrated in Figure~\ref{fig:thinking_layouts}. Each layout is
denoted by $a$-$b$-$c$-$d$, where the four numbers give the number of
thinking tokens inserted before the first to fourth target SID tokens.
For example, 2-2-0-0 places a pair of thinking tokens before $t^{y}_{1}$
and $t^{y}_{2}$, 0-0-2-2 and 0-0-0-2 move them to later SID positions,
and 2-2-2-2 inserts a pair before every SID token. These layouts only
change where thinking occurs, without view-specific temporal reading,
whereas \model{} inserts a single pair after the start token, with
$\bar{q}_{f}$ reading the full history and $\bar{q}_{r}$ reading the
last $w$ items. As shown in Table~\ref{tab:thinking_layouts}, no
positional layout performs best across all metrics: 2-2-0-0 outperforms
0-0-2-2 on five metrics, while 0-0-2-2 and 0-0-0-2 achieve the highest
H@20 and H@10 among positional layouts, respectively, and the denser
2-2-2-2 layout offers no consistent advantage. In contrast, \model{}
achieves the best results on all six metrics. These
results suggest that emphasizing finer-grained item information
or repeating thinking across semantic levels does not necessarily
provide better guidance for recommendation. Instead, \model{}
first integrates full-history and recent-history evidence into
a holistic preference context, providing a common foundation
for subsequent coarse-to-fine SID prediction. The comparison
supports the benefit of global preference synthesis before
generation over concentrating thinking on fine-grained SID.

\begin{table}[t]
\setlength\abovecaptionskip{0.2\baselineskip}
\setlength\belowcaptionskip{0.2\baselineskip}
    \centering
    \caption{Performance comparison of different thinking layouts and \model{}. H@$K$ and N@$K$ denote Hit Ratio and NDCG at $K$, respectively. The best results are boldfaced.}
    \label{tab:thinking_layouts}
    \begingroup
    \setlength{\tabcolsep}{3pt}
    \renewcommand{\arraystretch}{1.12}
    \footnotesize
    \resizebox{\columnwidth}{!}{%
    \begin{tabular}{@{}c|cccccc@{}}
        \toprule
        Layout & H@5 & H@10 & H@20 & N@5 & N@10 & N@20 \\
        \midrule
        2-2-0-0 & 0.0872 & 0.1079 & 0.1314 & 0.0740 & 0.0807 & 0.0866 \\
        \midrule
        0-0-2-2 & 0.0862 & 0.1071 & 0.1321 & 0.0732 & 0.0799 & 0.0863 \\
        0-0-0-2 & 0.0865 & 0.1081 & 0.1314 & 0.0736 & 0.0805 & 0.0865 \\
        \midrule
        2-2-2-2 & 0.0858 & 0.1073 & 0.1314 & 0.0734 & 0.0803 & 0.0864 \\
        \midrule
        \model{} & \textbf{0.0882} & \textbf{0.1088} & \textbf{0.1325} & \textbf{0.0751} & \textbf{0.0817} & \textbf{0.0877} \\
        \bottomrule
    \end{tabular}%
    }
    \endgroup
\end{table}

\definecolor{csNear}{HTML}{FCE8C8}
\definecolor{csMiss}{HTML}{F8D7D7}
\definecolor{csRecent}{HTML}{DCE8F8}
\definecolor{csCtx}{HTML}{EDEDED}
\definecolor{csHead}{HTML}{F4F4F4}
\definecolor{csNote}{HTML}{FFF6D6}

\providecommand{\cschip}[2]{%
  {\setlength{\fboxsep}{1.6pt}%
   \colorbox{#1}{\strut #2}}%
}
\providecommand{\csnear}[1]{\cschip{csNear}{#1}}
\providecommand{\csmiss}[1]{\cschip{csMiss}{#1}}
\providecommand{\csrec}[1]{\cschip{csRecent}{#1}}
\providecommand{\csctx}[1]{\cschip{csCtx}{#1}}
\providecommand{\caseimg}[1]{%
  \IfFileExists{#1}{%
    \includegraphics[height=1.25cm,keepaspectratio]{#1}\\[1pt]%
  }{}%
}

\begin{table*}[t]
\setlength{\abovecaptionskip}{0.2\baselineskip}
\setlength{\belowcaptionskip}{0.2\baselineskip}
\centering

\caption{Illustrative case-study scenarios on Instruments.
Histories are divided into earlier context
(\protect\csctx{gray}) and recent activity
(\protect\csrec{blue}).
Top-1 recommendations are highlighted as intent-consistent
but different from the target (\protect\csnear{orange}),
or inconsistent with the current activity
(\protect\csmiss{red}).
Rank denotes the position of the ground-truth item.}
\label{tab:case-study-illustration}

\begingroup
\footnotesize
\setlength{\tabcolsep}{3.5pt}
\renewcommand{\arraystretch}{1.25}

\newlength{\csRankW}
\setlength{\csRankW}{0.75cm}
\newlength{\csLabW}
\setlength{\csLabW}{1.45cm}
\newlength{\csItemW}
\setlength{\csItemW}{%
  \dimexpr
  (\textwidth-\csLabW-3\csRankW-14\tabcolsep)/3
  \relax
}
\newlength{\csSpanW}
\setlength{\csSpanW}{%
  \dimexpr\csItemW+\csRankW+2\tabcolsep\relax
}

\newcommand{\csspan}[1]{%
  \multicolumn{2}{>{\raggedright\arraybackslash}p{\csSpanW}}{#1}%
}
\newcommand{\csspanc}[1]{%
  \multicolumn{2}{>{\centering\arraybackslash}p{\csSpanW}}{#1}%
}

\begin{tabular}{%
@{}>{\raggedright\arraybackslash}p{\csLabW}%
>{\raggedright\arraybackslash}p{\csItemW}%
>{\centering\arraybackslash}p{\csRankW}%
>{\raggedright\arraybackslash}p{\csItemW}%
>{\centering\arraybackslash}p{\csRankW}%
>{\raggedright\arraybackslash}p{\csItemW}%
>{\centering\arraybackslash}p{\csRankW}@{}%
}
\toprule

\rowcolor{csHead}
 & \csspanc{\textbf{Case I: Recent intent shift}}
 & \csspanc{\textbf{Case II: Fine-grained variant}}
 & \csspanc{\textbf{Case III: Intent + variant}} \\

\cmidrule(lr){2-3}
\cmidrule(lr){4-5}
\cmidrule(l){6-7}

\textbf{Target}
 & \csspanc{%
     \caseimg{figures/case1_target.png}
     \textbf{Zildjian 5B nylon-tip drumsticks}}
 & \csspanc{%
     \caseimg{figures/case2_target.png}
     \textbf{Ernie Ball Not Even Slinky (.012--.056)}}
 & \csspanc{%
     \caseimg{figures/case3_target.png}
     \textbf{Dunlop Max-Grip nylon picks (.73\,mm)}} \\

\textbf{Earlier}
 & \csspan{%
     \csctx{Guitar strings}
     \csctx{Guitar straps}
     \csctx{Guitar tuners}}
 & \csspan{%
     \csctx{Electric-guitar accessories}
     \csctx{Amplifiers}}
 & \csspan{%
     \csctx{Broad guitar accessories}} \\

\textbf{Recent}
 & \csspan{%
     \csrec{E-drum kit}
     \csrec{Drum throne}
     \csrec{Zildjian sticks $\times$3}}
 & \csspan{%
     \csrec{Ernie Ball strings, multiple gauges}}
 & \csspan{%
     \csrec{Guitar strings}
     $\rightarrow$
     \csrec{Dunlop .73\,mm nylon picks}} \\

\midrule

\textbf{Method}
 & \textit{Top-1} & \textit{Rank}
 & \textit{Top-1} & \textit{Rank}
 & \textit{Top-1} & \textit{Rank} \\

TIGER
 & \csmiss{Guitar tuner}
 & $>20$
 & \csnear{Super Slinky (.009--.042)}
 & 6
 & \csmiss{Clip-on tuner}
 & $>20$ \\

w/o MVT
 & \csmiss{Guitar strings}
 & $>20$
 & \csnear{Regular Slinky (.010--.046)}
 & 3
 & \csmiss{Guitar-string set}
 & $>20$ \\

w/o HRA
 & \csnear{Zildjian 5A drumsticks}
 & 6
 & \csnear{Baritone (.013--.072)}
 & 8
 & \csnear{Dunlop picks (.88\,mm)}
 & 12 \\

\rowcolor{black!4}
\textbf{Ours}
 & \csnear{\textbf{Zildjian 5B wood-tip sticks}}
 & \textbf{3}
 & \csnear{\textbf{Beefy Slinky (.011--.054)}}
 & \textbf{2}
 & \csnear{\textbf{Dunlop standard nylon (.73\,mm)}}
 & \textbf{3} \\

\midrule

\textbf{Takeaway}
 & \csspan{%
     \cellcolor{csNote}
     \textbf{MVT} recovers the recent drum intent hidden by long-term guitar history.}
 & \csspan{%
     \cellcolor{csNote}
     \textbf{HRA} separates gauge-level variants within the correct category.}
 & \csspan{%
     \cellcolor{csNote}
     \textbf{MVT + HRA} are both needed for the new intent and exact thickness.} \\

\bottomrule
\end{tabular}
\endgroup
\end{table*}

\section{Effect of Path Organization}
\label{sec:path-organization}

To address the need for flexible context adaptation during SID
generation, we compare MV-Thinker with independently trained
single-path variants (Single-Path 1--3) and parallel paths with
temporal modeling (Parallel Paths) on Instruments, with TIGER as
the baseline. Table~\ref{tab:path-organization} shows that
Single-Path 1 performs best among the single-path variants on five
metrics, whereas Single-Path 2 leads on H@5, suggesting that no
single tested path is uniformly preferable. MV-Thinker surpasses
all single-path variants and Parallel Paths on all six metrics.
These gains show improved top-ranked relevance and ranking quality
across recommendation cutoffs compared with the tested isolated
or parallel alternatives. They suggest that organizing residual
refinements hierarchically helps MV-Thinker adapt user-conditioned
context more effectively during progressive generation.
By reusing shared transformations while supporting different
refinement depths, this design accommodates varying contextual
requirements across recommendation instances rather than imposing
a uniform adaptation on all users.

\begin{table}[t]
\setlength\abovecaptionskip{0.2\baselineskip}
\setlength\belowcaptionskip{0.2\baselineskip}
\centering
\caption{Recommendation performance of different path organization
strategies on Instruments. Single-Path 1--3 are trained separately;
Parallel Paths uses parallel paths with temporal modeling.
H@$K$ and N@$K$ denote Hit Ratio and NDCG at $K$, respectively.
The best results are boldfaced.}

\label{tab:path-organization}
\begingroup
\setlength{\tabcolsep}{3pt}
\renewcommand{\arraystretch}{1.08}
\footnotesize
\resizebox{\columnwidth}{!}{%
\begin{tabular}{@{}l|cccccc@{}}
\toprule
Method & H@5 & H@10 & H@20 & N@5 & N@10 & N@20 \\
\midrule
TIGER & 0.0853 & 0.1047 & 0.1293 & 0.0727 & 0.0790 & 0.0852 \\
\midrule
Single-Path 1 & 0.0867 & 0.1076 & 0.1316 & 0.0741 & 0.0807 & 0.0868 \\
Single-Path 2 & 0.0870 & 0.1058 & 0.1285 & 0.0733 & 0.0793 & 0.0851 \\
Single-Path 3 & 0.0861 & 0.1057 & 0.1307 & 0.0737 & 0.0800 & 0.0863 \\
Parallel Paths & 0.0876 & 0.1082 & 0.1320 & 0.0745 & 0.0812 & 0.0874 \\
\midrule
MV-Thinker & \textbf{0.0882} & \textbf{0.1088} & \textbf{0.1325} & \textbf{0.0751} & \textbf{0.0817} & \textbf{0.0877} \\
\bottomrule
\end{tabular}%
}
\endgroup
\end{table}

\section{Case Study}
\label{sec:case_study}

To examine how MVT and HRA influence intent recognition and
fine-grained item discrimination, we compare the top-1
recommendations and target ranks of \model{}, TIGER, and two
module ablations in three illustrative Instruments scenarios
(Table~\ref{tab:case-study-illustration}). Histories are divided
into earlier context and recent activity to relate recommendation
behavior to historical interests and current product requirements.
In Case~I, earlier interactions involve guitar strings, straps,
and tuners, whereas recent activity shifts toward drum equipment
and Zildjian drumsticks. TIGER and w/o MVT still recommend guitar
accessories and miss the target within the top~20. In contrast,
w/o HRA recommends Zildjian 5A drumsticks and ranks the target
5B nylon-tip drumsticks sixth, while \model{} recommends 5B
wood-tip sticks and ranks the target third. This contrast suggests
that MVT helps capture recent drum-related intent, while HRA
supports finer target ranking.
In Case~II, recent interactions concentrate on Ernie Ball strings
across multiple gauges. TIGER and w/o HRA recommend Super Slinky
(.009--.042) and Baritone (.013--.072), ranking the target Not Even
Slinky (.012--.056) sixth and eighth, respectively. w/o MVT
recommends Regular Slinky (.010--.046) and ranks the target third,
whereas \model{} recommends Beefy Slinky (.011--.054) and ranks
it second. All top-1 recommendations belong to the relevant
product family but differ in variant. The larger ranking decline
without HRA suggests that context refinement helps prioritize
the desired item beyond recognizing its general category.
In Case~III, recent activity shifts from guitar strings to Dunlop
.73-mm nylon picks. TIGER and w/o MVT recommend a tuner and a
string set, respectively, missing the target within the top~20.
w/o HRA recommends .88-mm picks and ranks the target Max-Grip
.73-mm picks twelfth. \model{} recommends standard .73-mm nylon
picks and ranks the target third, matching the desired thickness
but not the Max-Grip variant. This case suggests complementary
benefits from recent-intent recognition and attribute-level
refinement, while distinguishing better target ranking from
exact top-1 selection.
Overall, MVT helps interpret current activity within historical
context, while HRA supports finer item ranking through context
refinement. Their combination improves target ranking in these examples,
supporting recommendations that better reflect both the user's
current activity and specific product requirements.

\end{document}